%% file: ms.tex
\newcommand{\Tex}{\chi}
\newcommand{\dm}[1]{\textcolor{black}{#1}}
\newcommand{\dmr}[1]{\textcolor{black}{#1}}

\documentclass[journal]{elsarticle}

\usepackage{multirow}
\usepackage{subfigure}
\usepackage{epsfig}
\usepackage{graphicx}
\usepackage{amssymb}
\usepackage{amsmath}
\usepackage{color}
\usepackage{algorithm}
\usepackage{algorithmic}

\begin{document}

\begin{frontmatter}

\title{Consensus Based Medical Image Segmentation Using Semi-Supervised Learning And Graph Cuts}


\author{Dwarikanath Mahapatra}%
%

\address{IBM Research - Australia, Melbourne, Australia. \\
\textit{dwarim@au1.ibm.com}
}

\begin{abstract}
  Medical image segmentation requires consensus ground truth segmentations to be derived from multiple expert annotations. A novel approach is proposed that obtains consensus segmentations from experts using graph cuts (GC) and semi supervised learning (SSL). Popular approaches use iterative Expectation Maximization (EM) to estimate the final annotation and quantify annotator's performance. Such techniques pose the risk of getting trapped in local minima. We propose a self consistency (SC) score to quantify annotator consistency using low level image features. SSL is used to predict missing annotations by considering global features and local image consistency. The SC score also serves as the penalty cost in a second order Markov random field (MRF) cost function optimized using graph cuts to derive the final consensus label. Graph cut obtains a global maximum without an iterative procedure. Experimental results on synthetic images, real data of Crohn's disease patients and retinal images  show our final segmentation to be accurate and more consistent than competing methods. 
  \end{abstract}


\end{frontmatter}

\input{MultipleExperts_Intro}


\input{MultipleExperts_Method}

\input{MultipleExperts_Expts}

\input{MultipleExperts_Concl}


\section{References}


\bibliographystyle{elsarticle-num}
\bibliography{MultipleExperts_ref}

\end{document}

%% file: MultipleExperts_Intro.tex

\section{Introduction}
\label{sec:intro}

  Combining manual annotations from multiple experts is important in 
  medical image segmentation and computer aided diagnosis (CAD) tasks 
  such as performance evaluation of different 
  registration or segmentation algorithms, or  
  to assess the annotation quality of different
  raters through inter- and intra-expert variability \cite{LMS1}.
  Accuracy of the final (or consensus) segmentation determines to a large extent the accuracy of (semi-) automated segmentation and disease detection algorithms. 
  
\begin{figure}[t]
   \begin{tabular}{ccc}
	\includegraphics[height=1.5in,width=1.6in] {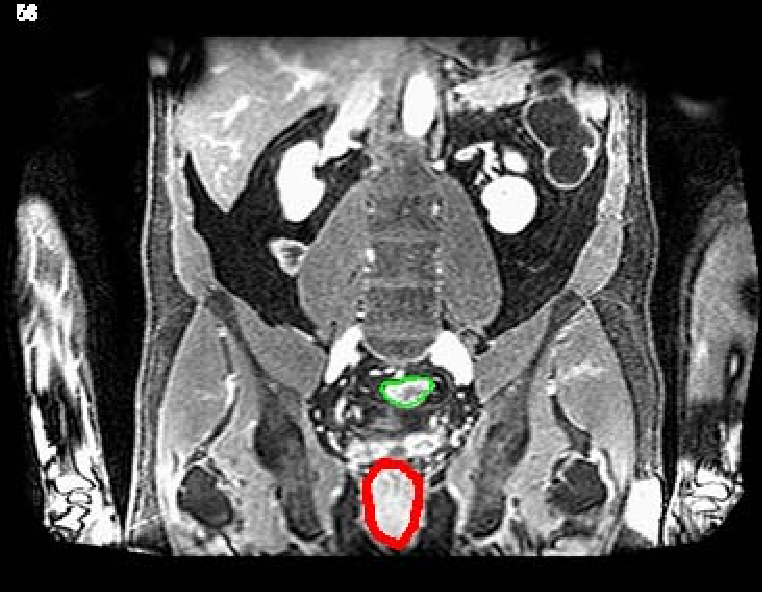} &
	  \includegraphics[height=1.5in,width=1.6in] {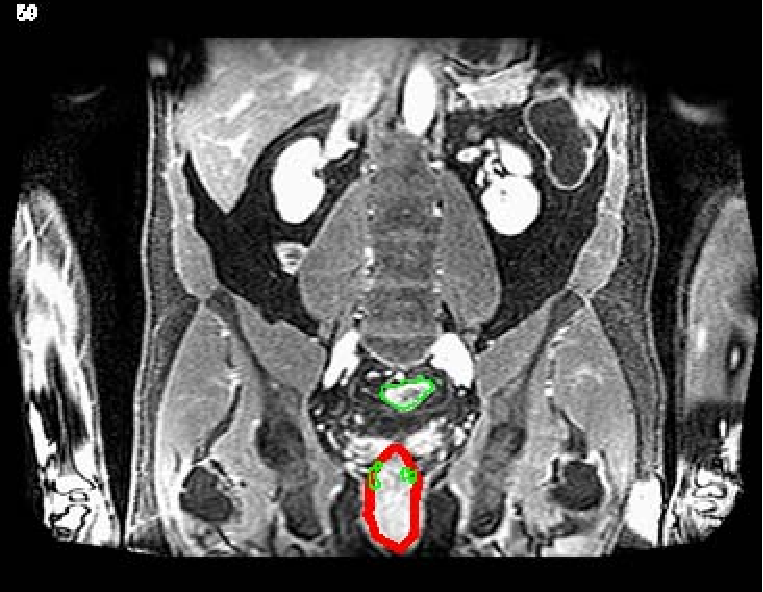}  &
	  \includegraphics[height=1.5in,width=1.6in] {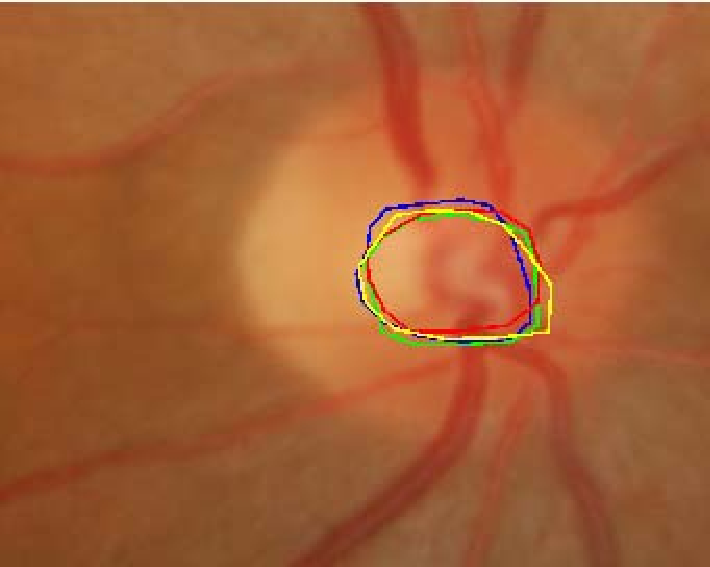}\\
	   (a) & (b) & (c) \\	
	  \end{tabular} 
	  \\                          
	%
	  %
	  \caption{(a)-(b) Illustration of subjectivity in annotating medical images. In both figures, red contour indicates diseased region as annotated by \textit{Expert~1} while green contour denotes diseased region as annotated by \textit{Expert~2}. (c) outline of optic cup by different experts.}
    \label{fig:ExpertAnnot1}
\end{figure}

  It is common for medical datasets to have annotations from different experts.
 Combining many experts' annotations is challenging due to their varying expertise levels,
 intra- and inter-expert variability, and missing labels of one or more experts.
 Poor consensus segmentations seriously affect the performance of segmentation algorithms,
 and robust fusion methods are crucial to their success. 
 In this work we propose to combine multiple expert annotations using semi-supervised learning (SSL) and graph cuts (GC). Its effectiveness is demonstrated on example annotations of Crohn's Disease (CD) patients on abdominal magnetic resonance (MR) images, retinal fundus images, \dm{and synthetic images}.
 Figure~\ref{fig:ExpertAnnot1} shows an example with two consecutive slices of a patient affected with CD. In both slices, the red contour indicates \dm{a} diseased region annotated by \textit{Expert~1} while green contour denotes diseased regions annotated by \textit{Expert~2}. Two significant observations can be made: 1) in Figure~\ref{fig:ExpertAnnot1} (a) there is no common region which is marked as diseased by both experts; 2) in Figure~\ref{fig:ExpertAnnot1} (b) the area agreed by both experts as diseased is very small. Figure~\ref{fig:ExpertAnnot1} (c) illustrates the challenges in retinal fundus images where different experts have different contours for the optical cup. The challenges of intra- and inter-expert variability are addressed by a novel 
 self-consistency (SC) score and the missing label information is predicted using SSL.

 \subsection{Related Work}

 Fusing expert annotations involves quantifying annotator performance.
 Global scores of segmentation quality for label fusion were proposed in \cite{LMS2,STAPLE}. 
  However, as suggested by Restif in \cite{LMS4} the computation of local performance is a better measure
  since it suits applications requiring varying accuracy in different image areas.
  Majority voting has also been used for fusing atlases of the brain in \cite{LMS7}. However, 
  it is limited by the use of a global metric for template
 selection which considers each voxel independently from
 others, and assumes equal contribution by each template
 to the final segmentation. It also produces 
locally inconsistent segmentations in regions of
high anatomical variability and poor registration.
  To address these limitations weighted majority voting was proposed in \cite{LMS10}
  that calculates weights based on intensity differences.
 This strategy  depends on intensity normalization and image
 registration and is error prone. 

 A widely used algorithm for label fusion is STAPLE \cite{STAPLE} that uses Expectation-Maximization (EM)
 to find sensitivity and specificity values maximizing the data likelihood. 
 These values quantify the quality of expert segmentations. 
 Their performance varies depending upon annotation accuracy, or anatomical 
 variability between templates \cite{LMS14}.
 Commowick et al. propose Local MAP STAPLE (LMSTAPLE) \cite{LocalMAPSTAPLE} that 
 addresses the limitations of STAPLE by using 
 sliding windows and Maximum A Posteriori (MAP) estimation, and defining a prior over 
 expert performance.
 Wang et al. \cite{WangPAMI13} exploit the correlation between different experts
 through a joint probabilistic model for 
 improved automatic brain segmentation. 
 Chatelain et al. in \cite{ChatelainMiccai13} use Random forests (RF) to determine 
 most coherent expert decisions with respect to the image by defining a consistency measure 
 based on information gain. They select the most relevant features to train the classifier, 
 and do not combine multiple expert labels. 
 %
 Statistical approaches such as COLLATE \cite{COLLATE} model the rating behavior of experts and use statistical analysis to 
 quantify their reliability. The final annotation is obtained using EM.
 The SIMPLE method combines atlas fusion and weight selection in an iterative procedure \cite{SIMPLE}.
 %
 Combining multiple atlases demonstrates the importance of anatomical information from multiple sources in segmentation tasks leading to  reduced error compared to a single training atlas \cite{SabuncuTMI2010,LotjonenNeuro2010}.


  \subsection{Our Contribution}

  The disadvantage of EM based methods is greater computation time, and the risk of 
  being trapped in local minimum. Consequently, the quantification of expert performance 
  might be prone to errors.
 Statistical methods such as \cite{STAPLER}
  require many simulated user studies to learn rater behavior, which may be biased 
  towards the simulated data. 

 Another common issue is missing annotation information from one or more experts. It is common practice to annotate only the interesting regions in medical images such as diseased regions or boundaries of an organ and disagreement between experts is a common occurrence. However in some cases we find that one or more experts do not provide any labels in some image slices, perhaps due to mistakes or inattention induced due to stress. In such cases it is important to infer the missing annotations and gather as much information as possible since it is bound to impact the quality of the consensus annotation. Methods like STAPLE predict missing labels that would maximize the assumed data likelihood function, which seems to be a strong assumption \dm{on the data distribution}. 

  Our work addresses the above limitations through the following \dm{contributions}:
  \begin{enumerate}
  \item SSL is used to predict missing annotation information. While SSL is a widely used concept in machine learning it has not been previously used to predict missing annotations. Such an approach 
  reduces the computation time since it predicts the labels in one step without
  any iterations as in EM based methods. By considering local pixel characteristics and global image information from the available labeled samples, SSL predicts missing annotations using global information but without making any strong assumptions of the form of the data generating function. 
  \item A SC score \dm{based on image features that best separate different training data} quantifies the reliability and accuracy of 
  each annotation. This includes both local and global
  information in quantifying segmentation quality.
  \item Graph cuts (GC) are used to obtain the final segmentation which
   gives a global optimum of the second order MRF cost function
  and also incorporates spatial constraints into the final solution.
  The SC is used to calculate the penalty costs for each possible class
  as reference model distributions cannot be defined in the absence of true label information. GC also pose minimal risk of being trapped in local minima compared to \dm{previous} EM based methods.
   \end{enumerate}
   We describe different aspects of our method in Sections \ref{met:feat}-\ref{sec:FinalLabel},
  present our results in Section~\ref{sec:expt} and
  conclude with Section~\ref{sec:disc}. 

%% file: MultipleExperts_Method.tex
  \input{MultipleExperts_ImageFeatures}

  \section{Learning Using Random Forests}
  \label{sec:method}
  
  Let us consider a multi-supervised learning scenario with a training set 
  $S=\{(x_n,y_n^{1},\cdots,y_n^{r})\}_{r=1}^{R}$ of samples $x_n$, and the 
  corresponding labels $y_n^{r}$ 
  provided by $R$ experts.
  A binary decision tree is a collection of nodes and leaves with each node containing a weak classifier that separates the data into two subsets
of lower entropy. 
Training a node $j$ on $S_j \subset S$ consists of finding the parameters
of the weak classifier that maximize the information gain ($IG_L$) of splitting labeled samples $S_j$
into $S_k$ and $S_l$:
  \begin{equation}
  IG_{j,L}(S_j,S_k,S_l) = H(S_j)-\frac{|S_k|}{S_j} H(S_k) -\frac{|S_l|}{S_j} H(S_l)
  \label{eqn:Il}
  \end{equation}
  where $H(S_i)$ is the empiric entropy of $S_i$, \dm{and $|.|$ denotes cardinality}. The parameters of the optimized weak classifier
 are stored in the node. Data splitting stops when we reach a predefined maximal depth, or when the
training subset does not contain enough samples. In this case, a leaf is created
that stores the empiric class posterior distribution estimated from this subset.

A collection of decorrelated decision trees increases generalization
power over individual trees.
Randomness is introduced by training each tree on a random subset of
the whole training set (bagging), and by optimizing each node over a random
subspace of the feature parameter space. At testing time, the output of the forest
is defined as the average of the probabilistic predictions of the $T$ trees. 
Note that the feature vector for every pixel consists of the features defined in Section~\ref{met:feat}.

  \subsection{Predicting Missing Labels}

%

  %
  Missing labels are commonly encountered when multiple experts annotate data. 
  We use semi-supervised learning (SSL)  to predict the missing labels.
  Unlike previous methods (\cite{BudvytisCVPR11}), a `single shot' RF method for SSL without the need for 
 iterative retraining was introduced in \cite{ForestBook}.
 We use this SSL classifier as it is shown to outperform other approaches \cite{ForestBook}.
%
  %
  %
 %
   For SSL the objective function encourages separation of the labeled training 
 data and simultaneously separates different high density regions. It is
achieved via the following mixed information gain for node $j$:
 \begin{equation}
IG_{j,SSL}=IG_{j,UL} + \alpha IG_{j,L}
\label{eqn:InfGain2}
\end{equation}
where $IG_{j,L}$ is defined in Eqn.~\ref{eqn:Il}. 
$I_{j,UL}$ depends on both labeled and unlabeled data, and is defined using differential entropies 
over continuous parameters as 
\begin{equation}
I_{j,UL}= \log |\Lambda(S_j)| - \sum_{i\in\{k,l\}} \frac{|S_j^{i}|}{|S_j|} \log |\Lambda(S_j)|
\label{eqn:DiffEnt}
\end{equation}
$\Lambda$ is the covariance matrix of the assumed multivariate distributions at each node.
For further details we refer the reader to \cite{ForestBook}. Thus the above cost function 
is able to combine the information gain from labeled and unlabeled data without the 
 need for an iterative procedure. 
 
  Each voxel has $r(\leq R)$ known labels and the 
 unknown $R-r$ labels are predicted by SSL.
 The feature vectors of all samples (labeled and unlabeled) are inputted to the 
 RF-SSL classifier which returns the missing labels. Note that although the same 
 sample (hence feature vector) has multiple labels, RF-SSL treats it as another sample
 with similar feature values.
 The missing labels are predicted based on 
 the split configuration (of decision trees in RFs) that  
 leads to maximal global information gain. 
 Hence the prediction of missing labels is not directly influenced
 by the other labels of the same sample
 but takes into account global label information \cite{ForestBook}.

   \section{Self Consistency of Experts}
   \label{sec:SC}
  
  Since the annotator is guided by
  visual features, \dm{such as intensity}, in distinguishing between different regions, 
  it is expected that for  reliable annotations 
  the region with a particular label would have consistent feature distributions. 
  Expert reliability is quantified by examining the information gain at different nodes while training a random forest on samples labeled by a particular expert. This helps us evaluate the consistency
of the experts with respect to the visual features. 
%
For each expert  $r$ we define an estimator $\widehat{E}^{r}_j$ of the
expectation of the information gain on the \emph{labeled training set} $S_j$ sent to node $j$ as
\begin{equation}
\widehat{E}^{r}_j=\frac{1}{\Theta_j} \sum_{\theta\in \Theta_j} {IG_{j,L}^{r}(S_j,S_k(\theta),S_l(\theta))}
\end{equation}
where $\Theta$ is a randomly selected subset of the feature parameters space. $\widehat{E}^{r}_j$ measures how well the data can be
separated according to the labels of each expert. However, it suffers from two
weaknesses in lower nodes of the tree: (i) it is evaluated from fewer samples, and
hence becomes less reliable, and (ii) it quantifies only the experts' local consistency,
without considering global consistency measures. Therefore, similar to \cite{ChatelainMiccai13} we 
define the performance level $q^{r}_j$ of each expert as a linear combination of the
estimators  $\widehat{E}^{r}_j$ from root to node $j$ as 
  \begin{equation}
  q_j^{r}=\frac{\sum_{d=0}^{D(j)} |S_d| \widehat{E}^{r}_{i_d(j)}}{\sum_{d=0}^{D(j)} |S_d|}.
  \end{equation}
  
  By weighting the estimators in proportion to the size of the training subset,
we give more importance to the global estimates of the experts' consistencies, but
still take into account their feature-specific performances. Once the parameters
$q^{r}_j$ have been computed, an expert's reliability or self consistency ($SC^{r}$) is calculated as the average performance level over all nodes $j$ in $T$ trees:
\begin{equation}
SC^{r}= \frac{\sum_j q_j^{r}}{T}
\label{eqn:SC}
\end{equation}
where $T$ is the total number of trees in the forest. Higher $SC^{r}$ indicates greater rater consistency.  
  To reduce computation time we select a region of interest (ROI) by taking the union of 
  all expert annotations and determining its bounding box rectangle. The size of the rectangle is 
  expanded by $\pm20$ pixels along rows and columns \dm{and $\pm2$ slices} to give the final ROI.

  \section{Obtaining Final Annotations}
  \label{sec:FinalLabel}

   The final annotation is obtained by optimising a second order MRF cost function that is given by, 
 
  \begin{equation}
E(L)=\sum_{s\in P} {D(L_s)} + \lambda \sum_{(s,t)\in N_s} {V(L_s,L_t)},
\label{eqn:MRF1}
\end{equation}
where $P$ denotes the set of pixels; $N_s$ is the $8$ neighbors of pixel $s$ (or sample $x$); $L_s$ is the label of $s$; $t$ is the neighbor of $s$, and $L$ is the set of labels for all $s$.  
$\lambda=0.06$ determines the relative contribution of penalty cost 
($D$) and smoothness cost ($V$). We have only $2$ labels ($L_s=1/0$ for object/background), although our 
method can also be applied to the multi-label scenario. The final labels are obtained by graph cut optimization
\dm{using Boykov's $\alpha-$expansion method. For details about the implementation we refer the reader to} \cite{BoykovFastApproximate}.

  The penalty cost for MRF is usually calculated with respect to a reference model of each class (e.g., distribution of intensity values). The implicit assumption is that the 
  annotators' labels are correct. 
  However, we aim to determine the \emph{actual labels} of each pixel and hence do not have access 
  to true class distributions. To overcome this problem we use the consistency  scores of experts to determine 
  the penalty costs for a voxel.
  Each voxel has $R$ labels (after predicting the missing labels). 
  Say for voxel $x$ the label $y^{r}$ (of the $r$th expert)
  is $1$, and the corresponding SC score is $SC_x^{r}$ (Eqn.\ref{eqn:SC}). Since SC is higher
  for better agreement with labels, the corresponding penalty cost for $L_x=1$ is 
  \begin{equation}
  D(L_x=1)^{r}=1-SC_x^{r},
  \end{equation} 
  where $L_x$ is the label of voxel $x$. Consequently, the corresponding penalty cost for label $0$ is 
  \begin{equation}
  D(L_x=0)^{r}=1-D(L_x=1)=SC_x^{r}.  
  \end{equation}
  %
  However, if the label $y^{r}$ (of the $r$th expert) is $0$, then the corresponding penalty costs are as follows
  \begin{equation}
  \begin{array}{ll}
  D(L_x=0)^{r}= & 1-SC_x^{r}   , \\
  D(L_x=1)^{r}= & 1-D(L_x=0)= SC_x^{r}.
  \end{array}  
  \end{equation}
   The individual penalty costs depend upon the labels given by the experts, while 
  the final penalty costs for each $L_x$ is the average of 
  costs from all experts,
  \begin{equation}
  \begin{array}{lr}
  D(L_x=1)=\frac{1}{R} \sum_{r=1}^{R} D(L_x=1)^{r}, \\
  D(L_x=0)=\frac{1}{R} \sum_{r=1}^{R} D(L_x=0)^{r}.
  \end{array}
  \end{equation}

   \textbf{Smoothness Cost (V)}:
   $V$ ensures a spatially smooth solution by penalizing discontinuities. \dm{We used a standard and popular formulation of the smoothness cost as originally proposed in \cite{BoykovFastApproximate}}. It is given by
%
 %
 %
\begin{equation}
  V(L_s,L_t) = \begin{cases}
     e^{-\frac{\left(I_s-I_t\right)^{2}}{2\sigma^{2}}}
  \cdot \frac{1}{\left\|s-t\right\|}, & L_s\neq L_t, \\
    0 & L_s=L_t. 
\end{cases}
\label{eqn:smoothseg}
\end{equation}
%
 %
 %
$I$ denotes the intensity.  Smoothness cost is
determined over a $8$ neighborhood system. 
%

%
%

%% file: MultipleExperts_ImageFeatures.tex
\section{Image Features}
\label{met:feat}

 Feature vectors derived for each voxel are used to predict any missing annotations from one or more experts. 
 Image intensities are normalized to lie between $[0,1]$.
Each voxel is described using intensity
statistics, texture and curvature entropy, and
spatial context features, and they are extracted from a $31\times31$ patch around each voxel. 
 In previous work \cite{MahapatraTMI_CD2013} we have used this same set of features to design a fully automated system for detecting and segmenting CD tissues from abdominal MRI. \dm{These patches were used on images of different sizes, $400\times400$ and $2896\times 1944$ pixels.} Through extensive experimental analysis of the RF based training procedure we identified context features to be most important followed by curvature, texture and intensity. Our hand crafted features also outperformed other feature combinations \cite{MahapatraABD12}. Since the current work focuses on a method to combine multiple expert annotations, we refer the reader to \cite{MahapatraTMI_CD2013} for details.

\subsection{Intensity Statistics}
\label{feat:int}

MR images commonly contain regions that do not form distinct spatial
patterns but differ in their higher order statistics \cite{Petrou1}.
Therefore, in addition to the features processed by the human visual
system (HVS), i.e., mean and variance, we extract  
skewness and kurtosis values from each voxel's neighborhood.

\subsection{Texture Entropy}
\label{feat:tex}

Texture maps are obtained from $2$-D Gabor filter banks
for each slice (at orientations
 $0^{\circ},45^{\circ},90^{\circ},135^{\circ}$ and scale $0.5,1$). 
They are partitioned into $9$ equal parts
corresponding to $9$ sectors of a circle. 
Figure~\ref{fig:illus} (a) shows the template for partitioning a patch into sectors
and extracting entropy features.
For each sector we calculate the texture entropy given by,
\begin{equation}
\Tex_{ani}^{r}=-\sum_{tex} p_{tex}^{r} \log p_{tex}^{r}.
\end{equation}
$p_{tex}^{r}$ denotes the probability distribution of texture
values in sector $r$. This procedure is repeated for all the $8$
texture maps over $4$ orientations and $2$ scales to extract a
($8\times9=$) $72$ dimensional feature vector.

\subsection{Curvature Entropy}
\label{met:shape}

  Different tissue classes have different curvature distributions and
  we exploit this characteristic for accurate discrimination between different tissue types.
Curvature maps are obtained from the gradient maps of the tangent along the $3$D surface.
The second fundamental form ($F2$) of these curvature maps is identical to the 
Weingarten mapping and the trace of the $F2$ matrix gives the mean curvature.
This mean curvature map is used for calculating curvature entropy. 
Details on curvature calculation are given in \cite{3dcurv,MahapatraTMI_CD2013}.
Similar to texture, 
curvature entropy is calculated from $9$ sectors of a patch
and is given by
\begin{equation} 
Curv_{ani}^{r}=-\sum_{\theta} p_{\theta}^{r} \log p_{\theta}^{r}.
\end{equation}
$p_{\theta}^{r}$ denotes the probability distribution of curvature
values in sector $r$, $\theta$ denotes the curvature values.
 Intensity, texture and curvature features
combined give a $85$ dimensional feature vector.

\dmr{
We use $2$D texture and curvature maps as the $3$D maps do not provide consistent features because of lower resolution in the $z$ direction compared to the $x$ and $y$ axis (voxel resolution was $1.02\times1.02\times2.0$ mm). Experimental results demonstrate that using $2$D features results in higher classification accuracy ($82\%$) in identifying diseased and normal samples when compared to using $3D$ features ($76\%$). We also resample the images using isotropic sampling and extract $3$D features, but the results are similar and favour the use of $2$D features. 
}

\subsection{Spatial Context Features:}
\label{met:context}

\begin{figure}[t]
   \begin{tabular}{cc}
	\includegraphics[height=1.0in,width=1.0in] {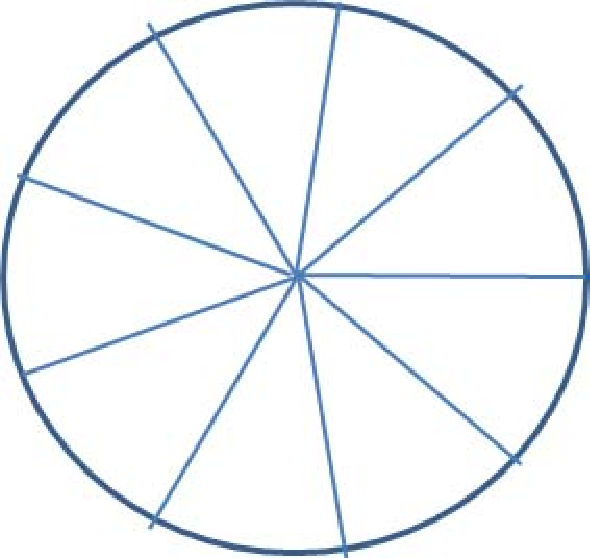} &
	  \includegraphics[height=1.0in,width=1.0in] {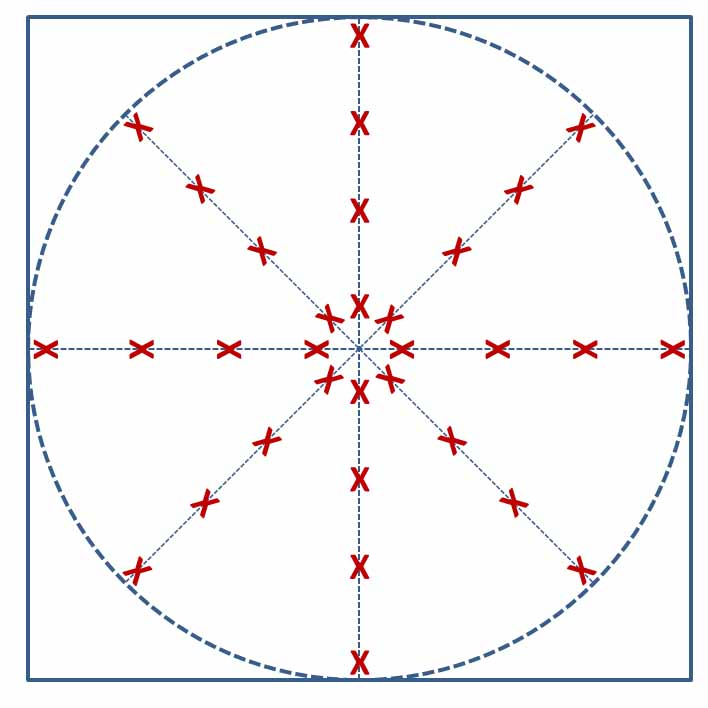}\\
	   (a) & (b) \\	
	  \end{tabular} 
	  \\                          
	%
	  %
	  \caption{(a) partitioning of patch for calculating anisotropy features; (b) template for 
	  calculating context features.}
    \label{fig:illus}
\end{figure}

 Context information is particularly important for 
 medical images because of 
 the regular arrangement of human organs \cite{Tu,ZhengSteerable}. 
Figure~\ref{fig:illus} (b) shows the template for context information
where the circle center is the current voxel and the
sampled points are identified by a red `X'.  At each point
corresponding to a `X' we extract a $3\times3$ region and
calculate the mean intensity, texture and curvature values. The
texture values were derived from the texture maps at
$90^{\circ}$ orientation and scale $1$. The `X's are located at
distances of $3,8,15,22$ pixels from the center, and the angle between
consecutive rays is $45^{\circ}$.  The values from the $32$ regions
are concatenated into a $96$ dimensional feature vector, and the final feature 
vector has $96+85=181$ values. \dmr{The choice of sampling distances and angles was determined experimentally on a small subset of images, with $3,8,15,22$ pixels and $45^{\circ}$ giving the best result in distinguishing between normal and diseased samples.}

%% file: MultipleExperts_Expts.tex
\section{Dataset Description}

\dm{We use real datasets from two different applications: $1)$ Crohn's disease detection, and $2)$ colour fundus retinal images originally intended for optic cup and disc segmentation, and a synthetic image dataset. Details of the different datasets are given below.}

\subsection{Crohn's Disease Dataset}

 For Crohn's Disease we use datasets from two different sources, one from the Academic
Medical Center ($AMC$), Amsterdam and the other from
University College of London Hospital ($UCL$). 

\begin{itemize}

\item $AMC$: The data was acquired from 
$25$ patients (mean age $38$ years, range, $25.6-59.6$ years, $15$ females) with luminal Crohn's disease that
had been approved by AMC's Medical Ethics Committee. All patients
had given informed consent to the prior study. 
Patients fasted four hours before a scan and drank $1600$ ml of
Mannitol ($2.5\%$) (Baxter, Utrecht, the Netherlands) one hour before
a scan. \dm{$T-1$ weighted} images were acquired using a
$3$-T MR imaging unit (Intera, Philips Healthcare, Best, The
Netherlands) with a $16$-channel torso phased array body coil. 
 The image resolution was $1.02$ mm $\times~1.02$
mm$\times~2$ mm/pixel, and the volume dimension was $400\times400\times100$
pixels.
%

%
\item $UCL$: Data from $25$ patients (mean age, $29.7$ years,
range, $17.4-54.3$ years, $12$ females) diagnosed with
small bowel Crohn's disease was used. \dm{$T-1$ weighted} images were acquired using a $3$T MR imaging
unit (Avanto; Siemens, Erlangen). 
The spatial resolution of the images was $1.02$ mm
$\times~1.02$ mm$\times~2$ mm per pixel. Two datasets have dimension of $512\times416\times48$, one
$512\times416\times64$, one $512\times512\times56$ and the rest 
$512\times512\times48$.  Ethical permission was given by the
University College London Hospital ethics committee, and informed
written consent was obtained from all participants.
%
\end{itemize}

\dmr{Each of the hospital MRI datasets was annotated by $4$ radiologists, two each from AMC and UCL.}
Consensus segmentations were obtained using 4 methods described in Section \ref{expt:real}.
The final segmentations of all $50$ patients are used to train 
a fully supervised method for detecting and segmenting CD tissues (details are 
given in Section \ref{expt:eval}) using $5-$fold cross validation.

\subsection{Colour fundus retinal images}

We use the DRISHTI-GS dataset \cite{Drishti} consisting of retinal fundus images from $50$ patients obtained using $30$ degree FOV at a resolution of $2896\times1944$ pixels. The optic cup and optic disc are manually segmented by $3$ ophthalmologists, and the consensus ground truth is also available. We choose this dataset because the final ground truth and annotations of individual experts are publicly available and facilitates accurate validation. 

\subsection{Evaluation Metrics}
\label{expt:eval}

 Availability of ground truth annotations makes it easier to evaluate the performance of 
 any segmentation algorithm. However, the purpose of our experiments is to estimate the
 actual ground truth annotations, and hence there is no direct method to estimate 
 the accuracy of the consensus annotations. 
 We adopt the following validation strategy using a fully supervised learning (FSL) framework:
 %
%
 
  %
  \begin{enumerate}
    \item Obtain the consensus segmentation from different methods.
  \item Train a separate RF classifier on the consensus segmentations of different methods in a $5-$fold cross validation setting. The same set of features as described in Section~\ref{met:feat} are used to describe each voxel. If the training labels were obtained using STAPLE then 
  the FSL segmentation of the test image is compared with the 
  ground truth segmentation from STAPLE only.
  \item Use the trained RF classifiers to generate probability maps for each voxel of the test image.
  \item Use the probability maps to obtain the final segmentation using the following 
  second order MRF cost function

\begin{equation}
   E(L)= \sum_{s\in P} {-\log \left(Pr(L_s)+ \epsilon \right)} +  \lambda \sum_{(s,t)\in N_s} {e^{-\frac{\left(I_s-I_t\right)^{2}}{2\sigma^{2}}} \cdot \frac{1}{\left\|s-t\right\|}},
\label{eqn:MRF2}
\end{equation}
%
%
where $Pr$ is the likelihood (from probability maps) 
previously obtained
using RF classifiers and $\epsilon=0.00001$ is a very small value to
ensure that the cost is a real number. The smoothness cost is same as Eqn~\ref{eqn:smoothseg}.

\item Obtain the final \emph{segmentation} using graph cuts. Note that this segmentation is part of the validation scheme and not for obtaining consensus annotations.
\end{enumerate}

This validation is similar to our previous method in \cite{MahapatraTMI_CD2013}, but 
 without using the supervoxels for region of interest detection. 
%
%
 The algorithm segmentations
 are compared with the `ground-truth' segmentations (the consensus segmentation obtained 
 by the particular method) using Dice Metric (DM) and Hausdorff distance (HD).
 Consensus segmentations with greater accuracy give better discriminative features
 and more accurate probability maps,
 and the classifiers obtained from these annotations can identify diseased regions more accurately.
 Thus we expect the resulting segmentations to be more accurate.
 The fusion method which most effectively combines the different annotations 
 is expected to give higher accuracy for the segmentations on the test data.

Dice Metric  measures the overlap between the segmented diseased region
obtained by our algorithm and reference manual annotations. It is
given by 
\begin{equation}
  DM = \frac{2 \left|A\cap M\right|}{\left| A\right| + \left| M\right|},
\end{equation}
where $A$ - segmentation from our algorithm and $M$ - manual annotations.
The $DM$ measure yields values between $0$ and $1$ where high DM
corresponds to a good segmentation. 

\textbf{Hausdorff Distance (HD)}: HD measures the distance between the contours
corresponding to different segmentations. If two curves are
represented as sets of points $A=\{a_1,a_2,\cdots.\}$ and
$M=\{m_1,m_2,\cdots.\}$, where each $a_i$ and $m_j$ is an ordered pair
of the $x$ and $y$ coordinates of a point on the curve, the distance
to the closest point (DCP) for $a_i$ to the curve $M$ is calculated.
The HD, defined as the maximum of the DCP's between the two curves, is:
\begin{equation}
\begin{array} {lr}
HD(A,M)= & \max( \max_i \{DCP(a_i,M)\}, \\
 & \max_j\{DCP(m_j,A)\}).
 \end{array}
\end{equation}

\dmr{The results between two different methods were compared using a paired $t-$test with a $5\%$ significance level that determines whether the two sets of results are statistically different or not. MATLAB's \textit{ttest2} function was used as it because it integrates better into our workflow and the result is returned as the $p-$value. Before performing the $t-$test we ensured that all essential assumptions are met namely, 1) all measurements are on a continuous scale; 2) the values are from a related group; 3) no significant outliers are present; 4) assumption of normality is not violated.}

Our whole pipeline was implemented in MATLAB on a $2.66$ GHz quad core
CPU running Windows $7$ with $4$ GB RAM.  The random forest code was a MATLAB
interface to the code in \cite{RFcode} written in the \emph{R} programming
language.The RF classifier had $50$ trees and its maximal tree depth was $20$.

\section{Experiments and Results}
\label{sec:expt}

  %
  %

	\subsection{Inter-expert  Agreement}
	\label{expt:inter}
	
	\dm{Each of the hospital MRI datasets was annotated by $4$ radiologists, two each from AMC and UCL.
	Thus each slice has $4$ different annotations and a mean annotation is calculated from them. The average DM between individual annotations and mean annotations was $91.5$ (minimum DM$=88.4$ and maximum DM$=94.3$). The corresponding average $p$ values from the paired $t-$test between the mean annotation of the individual annotations of that slice  was  $p=0.5241$ (minimum $p=0.1978$, maximum $p=0.5467$).
	The corresponding numbers for inter-expert agreement on retinal images was average DM$=94.3$ (minimum DM$=90.6$ and maximum DM$=96.0$), and average $p=0.6124$ (min $p=0.4582$, max $p=0.6715$). These values indicate good agreement between different experts. Since each expert annotated a slice only once we do not have the appropriate data to calculate intra-expert agreement.
	}

\subsection{MRF regularization strength $\lambda$ (Eqn.~\ref{eqn:MRF1})}
\label{expt:param}

To choose the MRF regularization strength $\lambda$ we choose a separate group of 
$7$ patient volumes (from both hospitals), and perform segmentation using our proposed
method but with $\lambda$ taking different values from $10$ to
$0.001$. The results are summarized in Table~\ref{tab:lambda}.  The
maximum average segmentation accuracy using Dice Metric (DM) was
obtained for $\lambda=0.06$ which was fixed for subsequent experiments. Note that these $7$ datasets
were a mix of patients from the two hospitals and 
\emph{not~ part} of the test dataset used for evaluating our algorithm.

 \begin{table}[t]
\centering
\caption{Change in segmentation accuracy with different values of  $\lambda$ (Eqn.~\ref{eqn:MRF1}). $DM$ is in $\%$.}
\begin{tabular}{|l|l|l|l|l|l|l|l|l|l|}
\hline
{$\lambda$} & {$10$}  & {$5$} & {$1$} & {$0.5$} & {$0.1$} & {$0.06$} & {$0.02$} & {$0.01$} & {$0.001$}  \\ \hline
{DM} & {71.4} & {72.8} & {75.4} & {80.2} & {82.8}  & {88.7} & {87.2} & {87.4} & {86.1} \\   \hline
%
\end{tabular}
\label{tab:lambda}
\end{table}

\subsection{Influence of Number of Trees}

The effect of the number of trees ($N_T$) on the segmentation is evaluated by 
varying them and observing the final segmentation accuracy (DM values)
 on the $7$ datasets mentioned above.
The results are summarized in  Table~\ref{tab:trees}.
%
For $N_T>50$ there is no
significant increase in DM ($p>0.41$) but the training time increases
significantly.  The best trade-off between $N_T$ and DM is achieved
for $50$ trees and is the reason behind our choice in the RF
ensemble.
The tree depth was fixed at $20$ after cross validation comparing tree depth, 
and resulting classification accuracy.

\begin{table}[t]
\centering
\caption{Effect of number of trees in RF classifiers ($N_T$) on segmentation accuracy and training time ($T_{Tr}$) of $RF-SSL$. $DM$ is in $\%$.} 
\begin{tabular}{|l|l|l|l|l|l|l|l|l|}
\hline
{$N_T$} & {$5$}  & {$7$} & {$10$} & {$20$} & {$50$} & {$70$} & {$100$} & {$150$} \\ \hline
{DM} & {82.5} & {84.7} & {86.6} & {88.3} & {91.7} &  {91.8} & {91.7} & {91.7}  \\   \hline
{$T_{Tr}$} & {0.20T} & {0.21T} & {0.42T} & {0.8T} & {T} & {1.4T} & {2.2T} & {3.4T} \\ \hline
\end{tabular}
\label{tab:trees}
\end{table}

\input{MultipleExperts_Expts_syn}

  \subsection{Real Patient Crohn's Disease Dataset}
  \label{expt:real}

  For the CD patient datasets we show consensus segmentation results for $GC_{ME-All}$, $GC_{ME}$, $GC_{ME-wSSL}$
 COLLATE, Majority Voting (MV), and LMStaple.
  Although, all the $4$ experts annotated every image, in order to test 
  our SSL based prediction strategy, we intentionally removed $1$ or $2$ annotations
  for each image/volume slice.

\begin{figure*}[t]
\begin{minipage}[b]{0.315\linewidth}
\centerline{\epsfig{figure=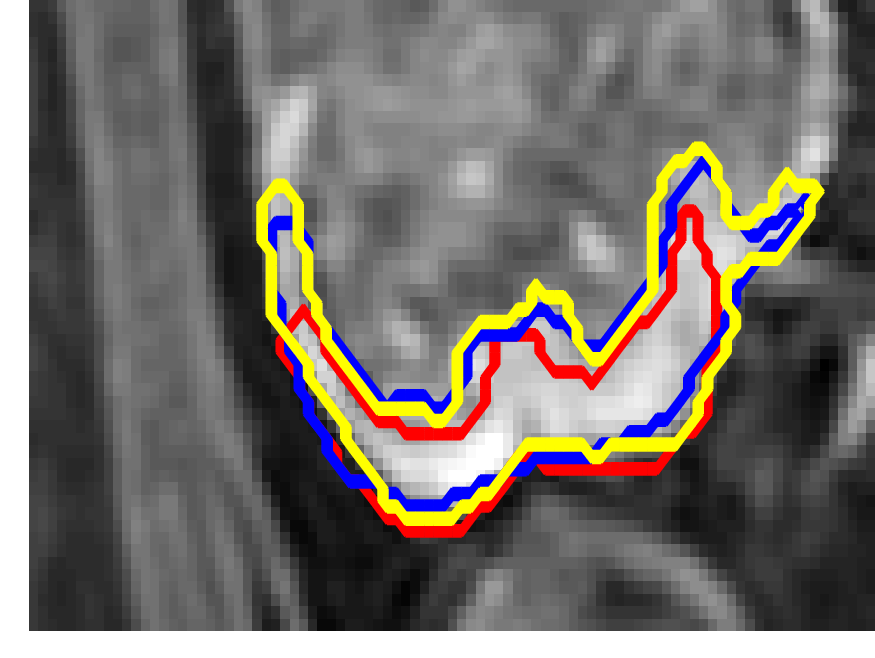,height=4cm,width=4cm}}  
\centerline{(a)}\medskip   \vspace{-0.25cm}
\end{minipage}
\hfill
\begin{minipage}[b]{0.315\linewidth}
\centerline{\epsfig{figure=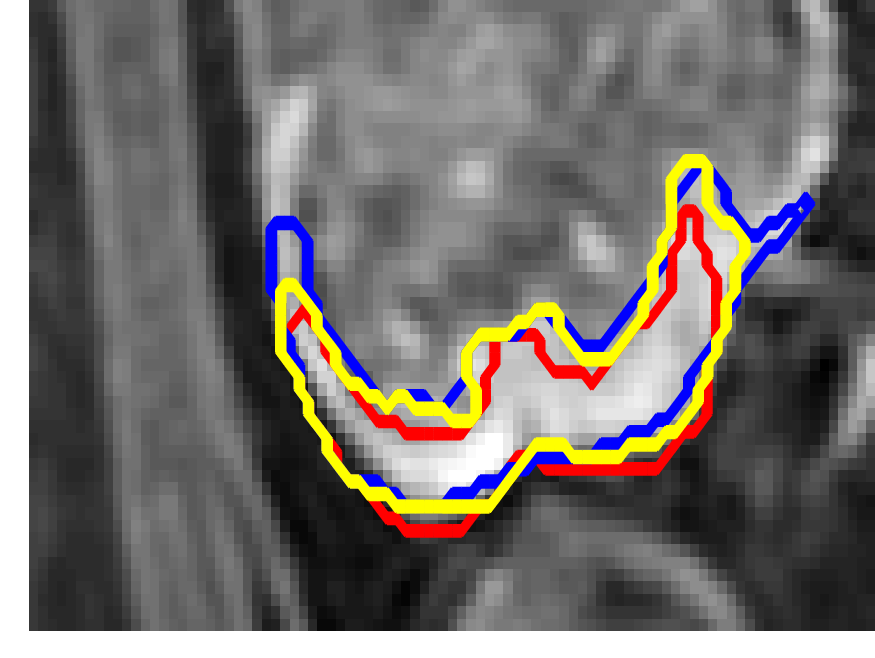,height=4cm,width=4cm}}  
\centerline{(b)}\medskip   \vspace{-0.25cm}
\end{minipage}
\hfill
\begin{minipage}[b]{0.315\linewidth}
\centerline{\epsfig{figure=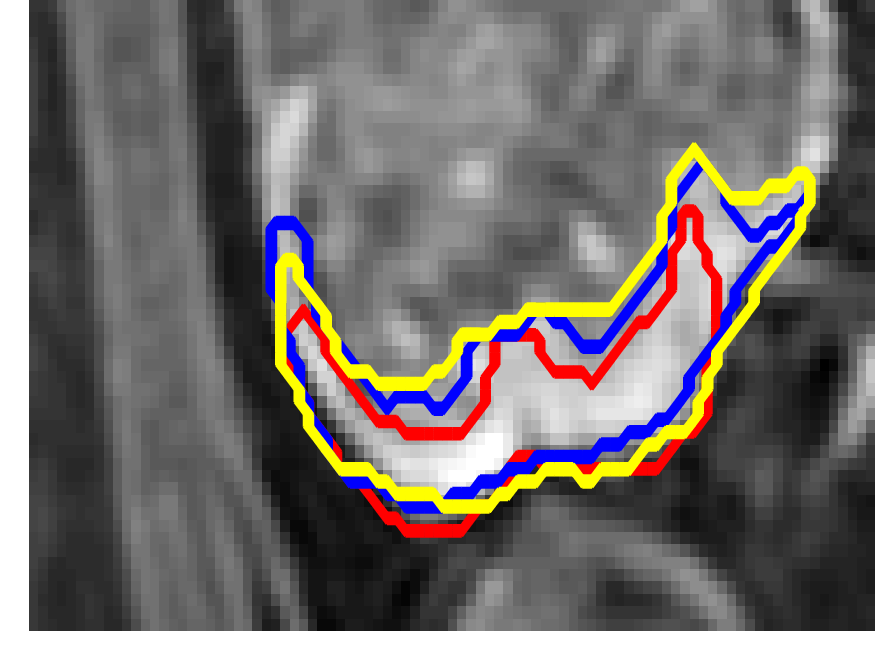,height=4cm,width=4cm}}  
\centerline{(c)}\medskip   \vspace{-0.25cm}
\end{minipage}
\hfill
\begin{minipage}[b]{0.315\linewidth}
\centerline{\epsfig{figure=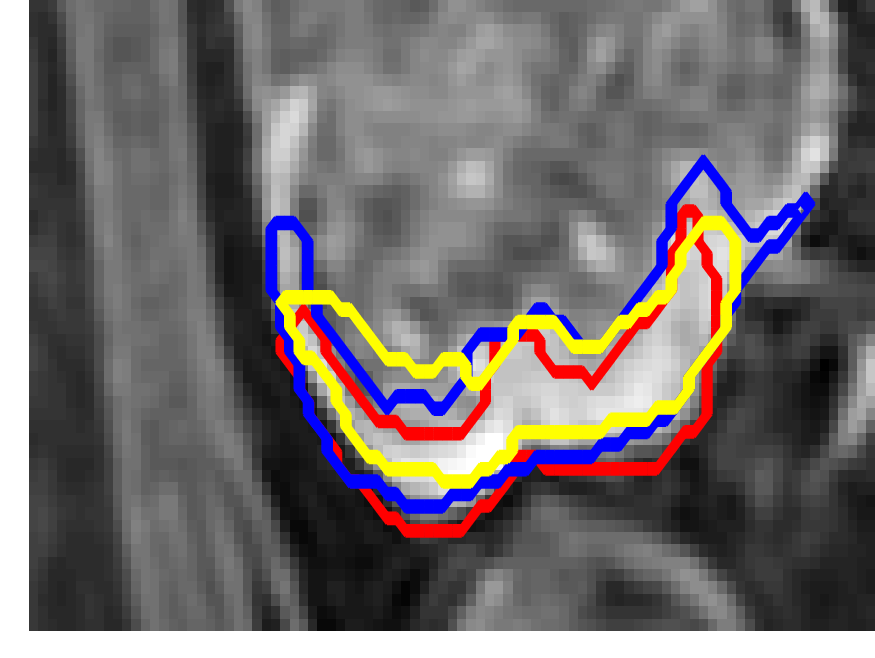,height=4cm,width=4cm}}  
\centerline{(d)}\medskip   \vspace{-0.25cm}
\end{minipage}
\hfill
\begin{minipage}[b]{0.315\linewidth}
\centerline{\epsfig{figure=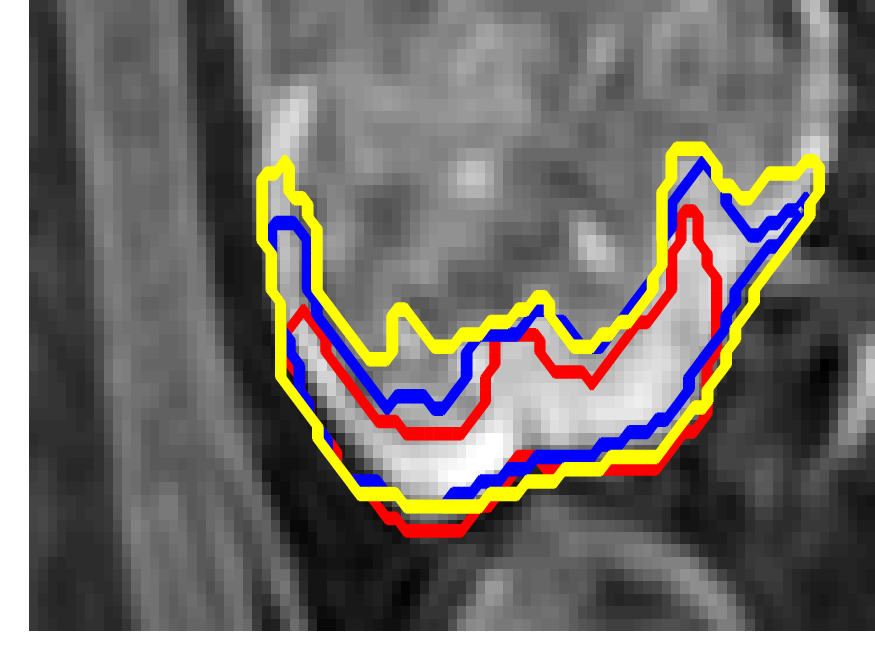,height=4cm,width=4cm}}  
\centerline{(e)}\medskip   \vspace{-0.25cm}
\end{minipage}
\hfill
\begin{minipage}[b]{0.315\linewidth}
\centerline{\epsfig{figure=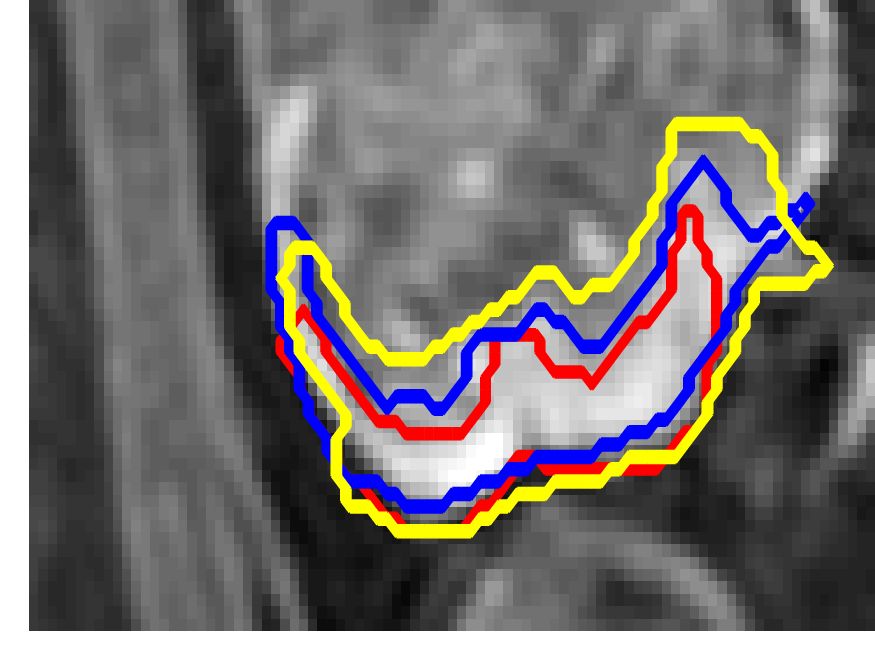,height=4cm,width=4cm}}  
\centerline{(f)}\medskip   \vspace{-0.25cm}
\end{minipage}
\hfill
\vspace{-0.4cm}
\caption{The predicted ground truth for UCL Patient 23 by different methods: (a) $GC_{ME-All}$; (b) $GC_{ME}$; (c) \cite{LocalMAPSTAPLE}; (d) \cite{COLLATE}; (e) \cite{LMS7}; and (f) $GC_{ME-wSSL}$. Red and blue contours are expert annotations and yellow is the final annotation obtained by the respective methods.}
\label{fig:GroundTruths_CD}
\end{figure*}

  Figure~\ref{fig:GroundTruths_CD} shows the predicted ground truth for $6$
  fusion strategies using only two expert labels. We show results for two experts 
  due to the ease in showing the different annotations in one image. Displaying three or more 
  expert annotations with the consensus segmentation makes the images very crowded and 
  hence difficult to interpret. Since our purpose is to show the relative merit of different methods,
  two expert annotations also serve the same purpose.
   
\begin{figure*}[t]
\begin{minipage}[b]{0.315\linewidth}
\centerline{\epsfig{figure=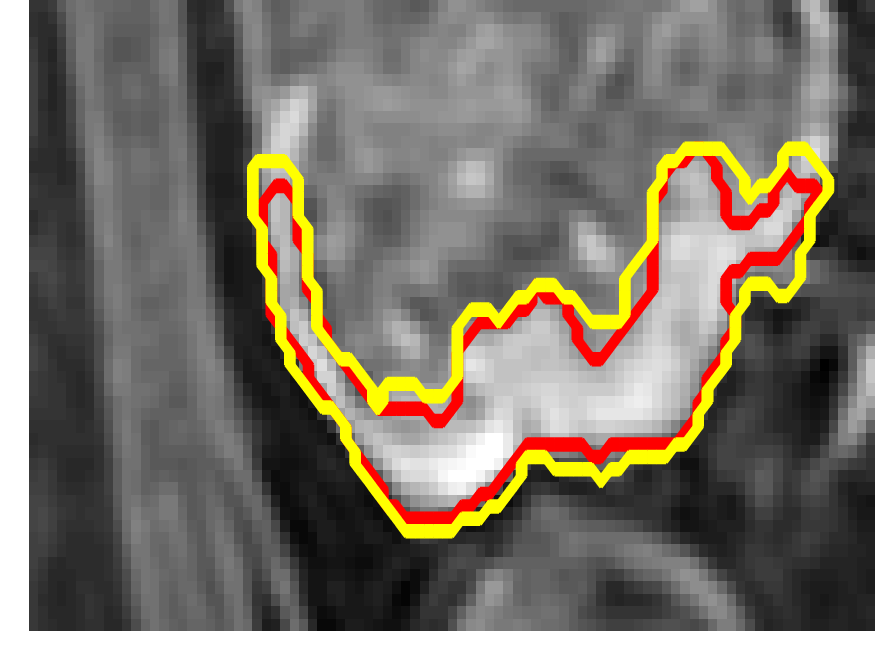,height=4cm,width=4cm}}  
\centerline{(a)}\medskip   \vspace{-0.25cm}
\end{minipage}
\hfill
\begin{minipage}[b]{0.315\linewidth}
\centerline{\epsfig{figure=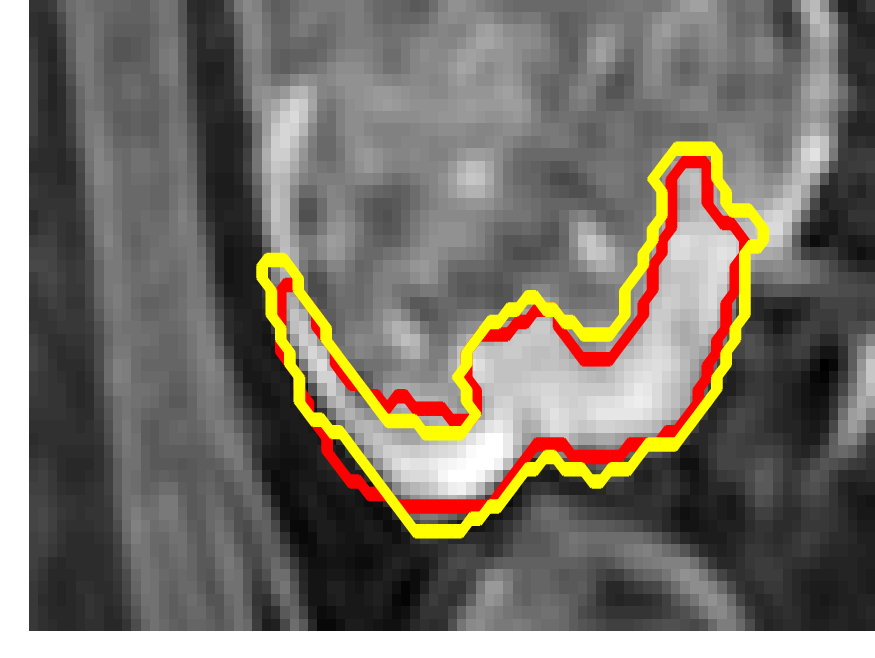,height=4cm,width=4cm}}  
\centerline{(b)}\medskip   \vspace{-0.25cm}
\end{minipage}
\hfill
\begin{minipage}[b]{0.315\linewidth}
\centerline{\epsfig{figure=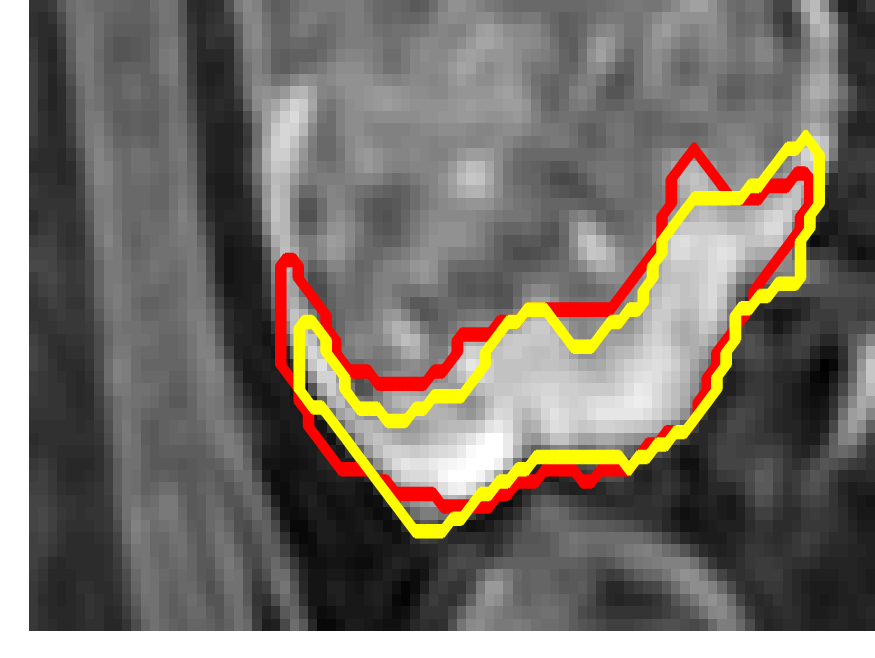,height=4cm,width=4cm}}  
\centerline{(c)}\medskip   \vspace{-0.25cm}
\end{minipage}
\hfill
\begin{minipage}[b]{0.315\linewidth}
\centerline{\epsfig{figure=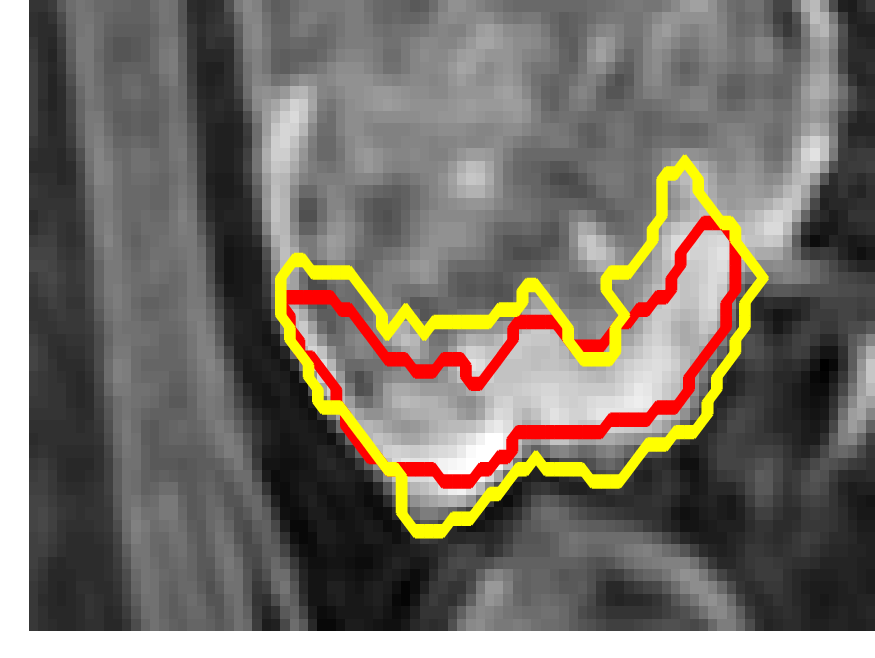,height=4cm,width=4cm}}  
\centerline{(d)}\medskip   \vspace{-0.25cm}
\end{minipage}
\hfill
\begin{minipage}[b]{0.315\linewidth}
\centerline{\epsfig{figure=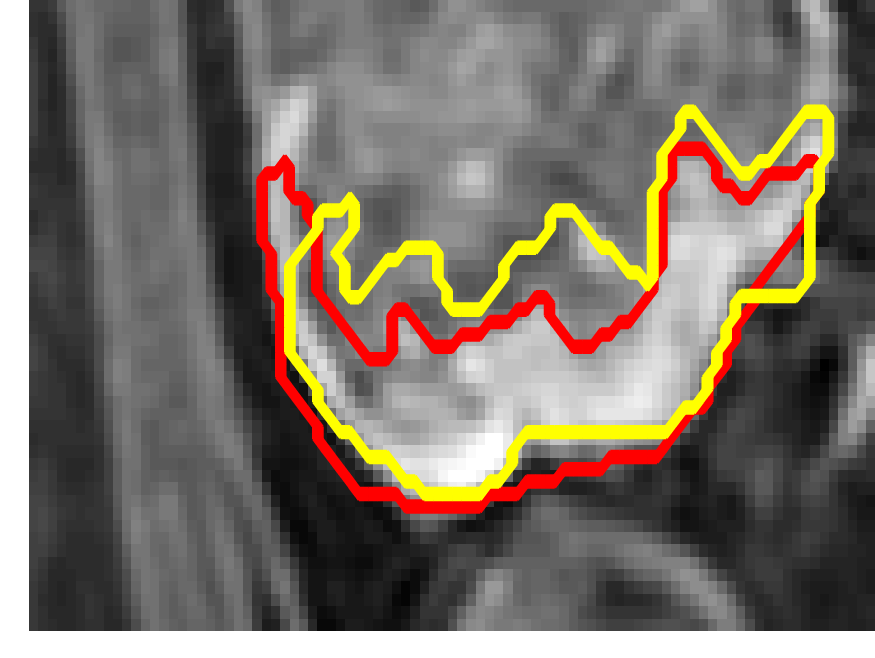,height=4cm,width=4cm}}  
\centerline{(e)}\medskip   \vspace{-0.25cm}
\end{minipage}
\hfill
\begin{minipage}[b]{0.315\linewidth}
\centerline{\epsfig{figure=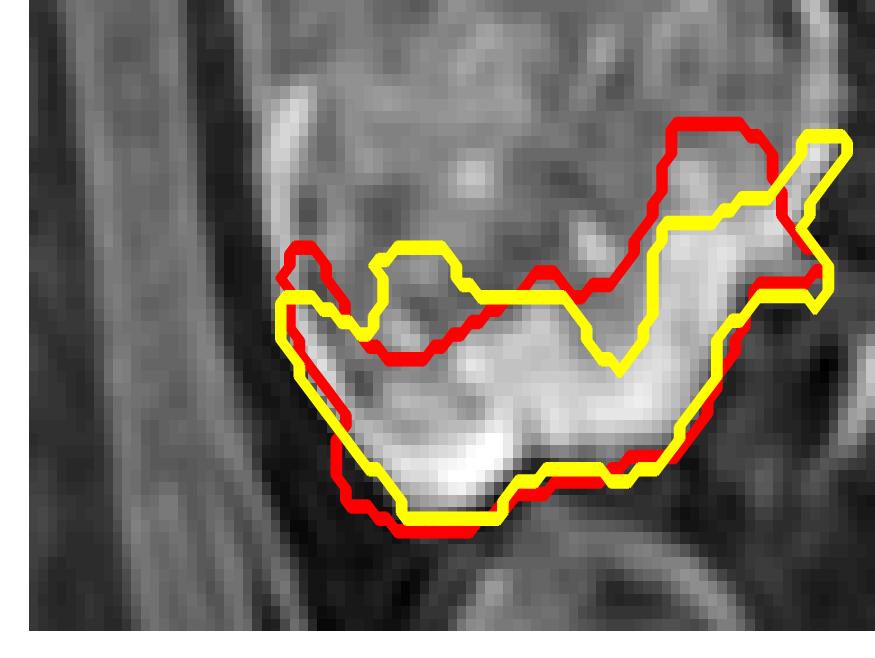,height=4cm,width=4cm}}  
\centerline{(f)}\medskip   \vspace{-0.25cm}
\end{minipage}
\hfill
\vspace{-0.4cm}
\caption{Segmentation results on UCL patient 23 for: (a) $GC_{ME-All}$; (b) $GC_{ME}$; (c) \cite{LocalMAPSTAPLE}; (d) \cite{COLLATE}; (e) \cite{LMS7}; and (f) (a) $GC_{ME-wSSL}$;. Red contour is the corresponding ground truth generated by the fusion method, and yellow contour is the algorithm segmentation obtained as described in Section~\ref{expt:eval}.}
\label{fig:SegRes1}
\end{figure*}

\begin{figure*}[t]
\begin{minipage}[b]{0.315\linewidth}
\centerline{\epsfig{figure=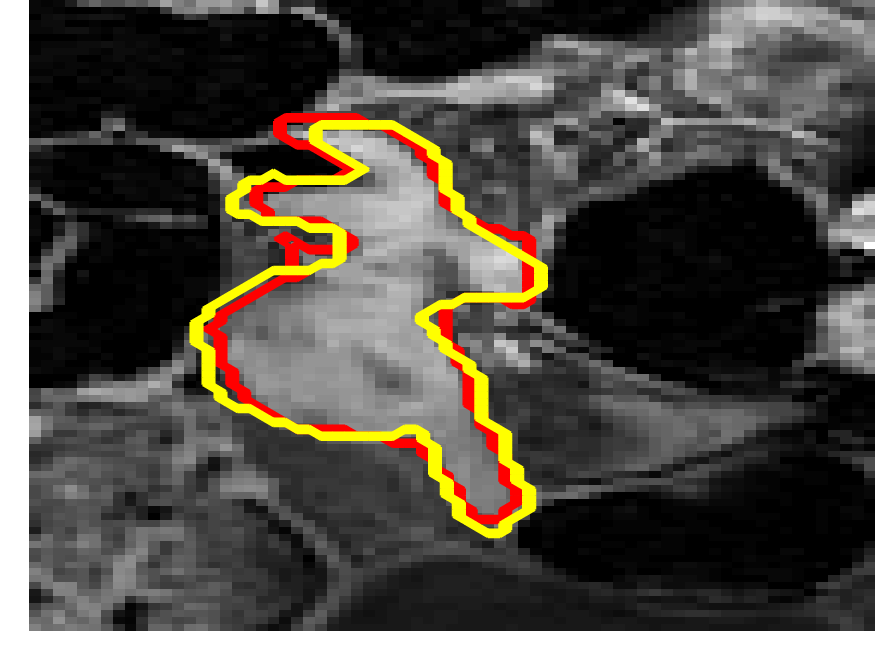,height=4cm,width=4cm}}  
\centerline{(a)}\medskip   \vspace{-0.25cm}
\end{minipage}
\hfill
\begin{minipage}[b]{0.315\linewidth}
\centerline{\epsfig{figure=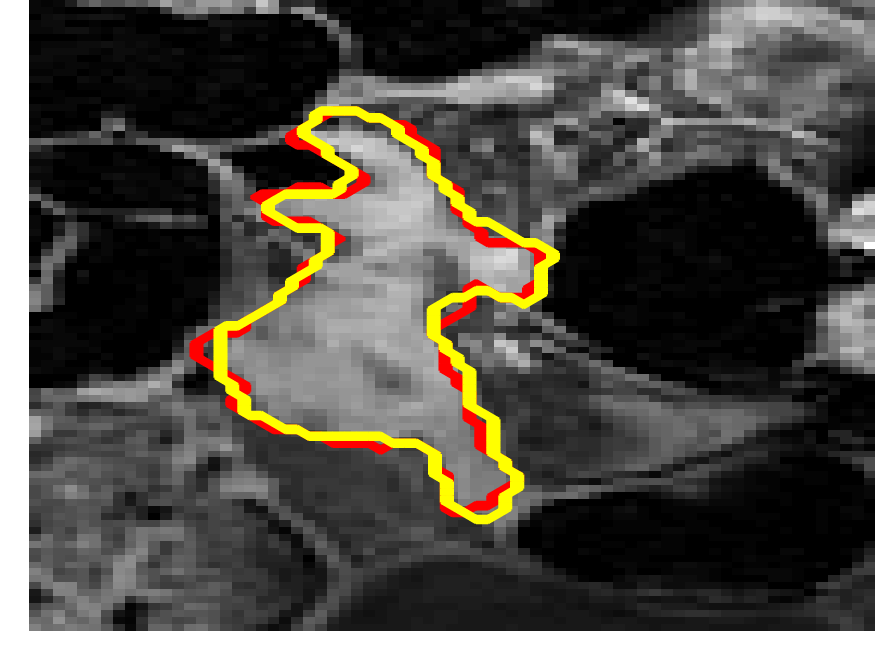,height=4cm,width=4cm}}  
\centerline{(b)}\medskip   \vspace{-0.25cm}
\end{minipage}
\hfill
\begin{minipage}[b]{0.315\linewidth}
\centerline{\epsfig{figure=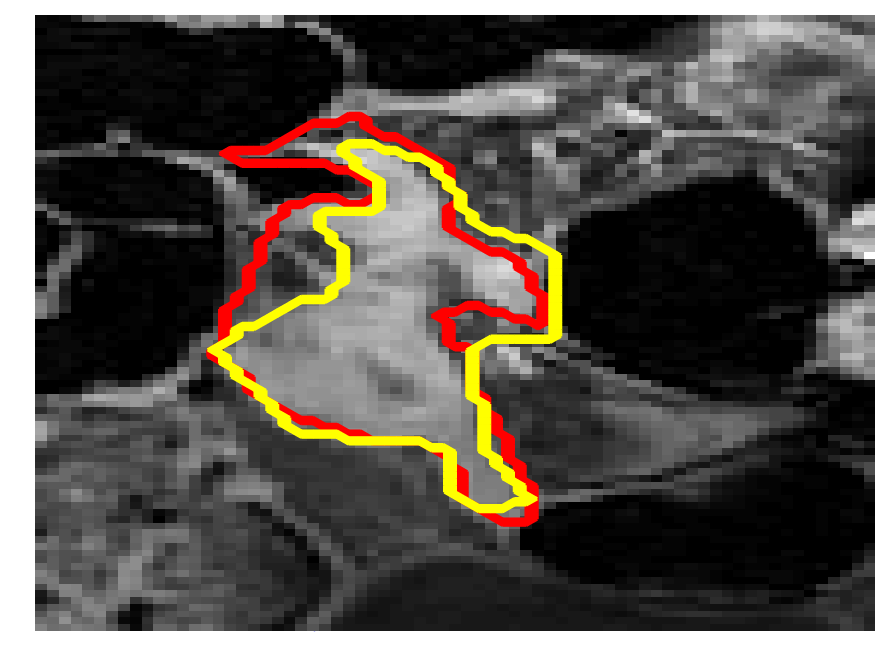,height=4cm,width=4cm}}  
\centerline{(c)}\medskip   \vspace{-0.25cm}
\end{minipage}
\hfill
\begin{minipage}[b]{0.315\linewidth}
\centerline{\epsfig{figure=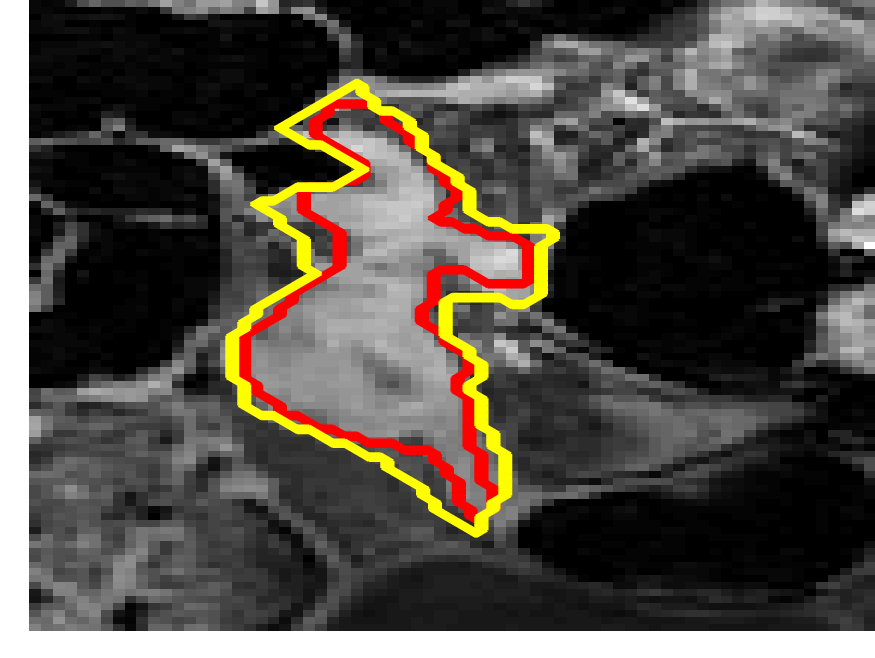,height=4cm,width=4cm}}  
\centerline{(d)}\medskip   \vspace{-0.25cm}
\end{minipage}
\hfill
\begin{minipage}[b]{0.315\linewidth}
\centerline{\epsfig{figure=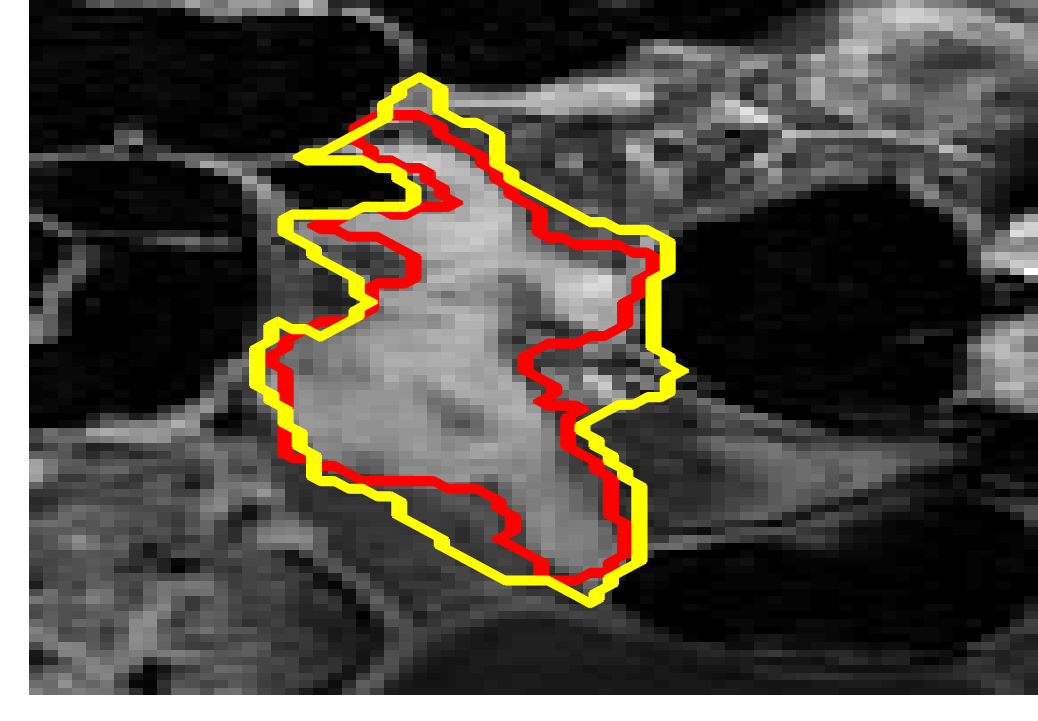,height=4cm,width=4cm}}  
\centerline{(e)}\medskip   \vspace{-0.25cm}
\end{minipage}
\hfill
\begin{minipage}[b]{0.315\linewidth}
\centerline{\epsfig{figure=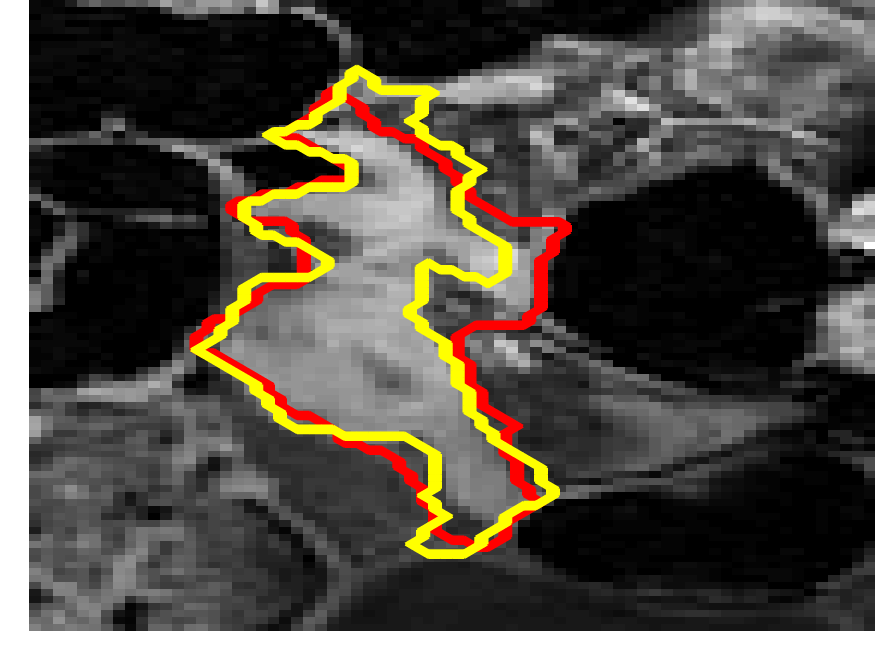,height=4cm,width=4cm}}  
\centerline{(f)}\medskip   \vspace{-0.25cm}
\end{minipage}
\hfill
\vspace{-0.4cm}
\caption{Segmentation results on AMC patient 15 for: (a) $GC_{ME-All}$; (b) $GC_{ME}$; (c) \cite{LocalMAPSTAPLE}; (d) \cite{COLLATE}; (e) \cite{LMS7}; and (f) (a) $GC_{ME-wSSL}$;. Red contour is the corresponding ground truth generated by the fusion method, and yellow contour is the algorithm segmentation obtained as described in Section~\ref{expt:eval}.}
\label{fig:SegRes2}
\end{figure*}

   Figures~\ref{fig:SegRes1},\ref{fig:SegRes2} show segmentation results for two 
   patients ($UCL$ Patient 23 and $AMC$ Patient 15) using all the 
   $6$ fusion strategies mentioned above and 
     Table~\ref{tab:SegRes1} summarizes their average performance 
     over all $50$ patients.  
  From the visual results and quantitative measures it is clear that 
  $GC_{ME-All}$ gives the highest DM and lowest HD values, followed by 
  $GC_{ME}$, \cite{LocalMAPSTAPLE}, \cite{COLLATE}, \cite{LMS7}, and
  $GC_{ME-wSSL}$.
  Since $GC_{ME-All}$ had access to all annotations, it obviously performed best.
  However $GC_{ME}$'s performance is very close and a Student $t-$test with $GC_{ME-All}$
  gives $p<0.042$ indicating very small difference in the two results. 
  Thus we can effectively conclude that $GC_{ME}$ does a very good job in predicting missing
  annotations.
  Importantly, $GC_{ME}$ performs much better than all other methods ($p<0.01$).
  The results show SSL effectively predicts missing annotation information since 
  $GC_{ME-wSSL}$ shows a significant drop in performance from $GC_{ME}$ ($p<0.01$).

   If the consensus segmentation is inaccurate then the subsequent training is also 
   flawed because the classifier learns features from many voxels whose label is inaccurate.
   As a result, in many cases the final segmentation includes regions which do not exhibit 
   any disease characteristics as confirmed by our medical experts. 
  Another limitation of sub-optimal label fusion is the wide variation in 
   segmentation performance of that particular method. The standard deviation of 
   \cite{LMS7} is much higher than $GC_{ME}$ indicating inconsistent segmentation quality.
   A good fusion algorithm should assign lower reliability scores to inconsistent segmentations,
   which is achieved by $GC_{ME}$ as is 
   evident from the low variation in its DM scores.

   An important factor limiting 
   the performance of LMStaple is its prediction of sensitivity and specificity 
   parameters from the annotations without considering their overall consistency. 
   Our SC score takes into account both global and local information
   and is able to accurately quantify a rater's consistency.
   The effect of SC is also highlighted through experiments on synthetic images (Section~\ref{expt:synth})
   Secondly, LMStaple may be prone to being trapped in local minimum due to the 
   iterative EM approach. On the contrary, we employ graph cuts which is almost 
   always guaranteed to give a global minimum. This makes the final output (the consensus segmentation)
   much more accurate and robust. 
   COLLATE also suffers due to its reliance on an EM based approach.

  \begin{table}[t]
	\centering
		\caption{Quantitative measures for segmentation accuracy on CD images. DM- Dice Metric in $\%$; HD is Hausdorff distance in mm and $p$ is the result of Student $t-$tests with respect to $GC_{ME}$.}
		\begin{tabular}{|l|l|l|l|l|l|l|}
		\hline
      {}  & {$GC$} & {$GC_{ME}$} & {LMSTAPLE} & {COLLATE} & {MV} & {$GC$}  \\ 
      {}  & {$_{ME-All}$} & {} & {\cite{LocalMAPSTAPLE}}  & {\cite{COLLATE}} & {\cite{LMS7}} & {$_{ME-wSSL}$}  \\ \hline 
      {DM} & {92.6$\pm$2.4} & {91.7$\pm$3.0} & {87.3$\pm$4.5} & {85.1$\pm$5.3} & {83.8$\pm$7.3} & {82.3$\pm$9.0} \\   \hline
      {HD} & {7.4$\pm$2.6} &  {8.2$\pm$3.3} & {9.8$\pm$4.8} &  {12.0$\pm$6.2} & {13.9$\pm$7.4} & {14.7$\pm$8.2}  \\   \hline
      {$p$} & {$0.042$} & {-} & {$<0.01$} & {$<0.01$} & {$<0.01$} & {$<0.01$}\\ \hline 
      %
	\end{tabular}
\label{tab:SegRes1}
\end{table}

\input{MultipleExperts_Expts_Retina}

\vspace{.5cm}

\subsection{Computation Time}

 Since the size of the annotations varies depending 
 on the diseased area (ROI varies between $70\times80$ to $170\times200$), 
 an average fusion time for an annotation may be misleading.
 Therefore we calculate an average fusion time per pixel, which is the highest for 
  LMStaple at $0.3$ seconds followed by 
 COLLATE ($0.22$ seconds), $GC_{ME}$ ($0.1$ seconds) and majority (voting) MV ($0.05$ seconds). Other variations of 
 $GC_{ME}$ take almost the same time as $GC_{ME}$. 
 Note that we report only the time for fusing the annotations and
 not the total segmentation time as the segmentation time is 
 the same for all cases since a RF based framework is used.
 The segmentation time is an additional $0.2$ seconds per pixel.

 These results clearly show the faster performance by our method due to employing
 SSL and GC for predicting missing annotations and obtaining the final annotation.
 The EM based LMStaple algorithm is nearly $3$ times slower than $GC_{ME}$,
 while COLLATE is $2$ times slower because of many computations. 
 Majority voting is faster than all other methods
 because of its simple approach to predicting final annotations. However, its performance is
 the worst.

%% file: MultipleExperts_Expts_syn.tex
 
 \subsection{Synthetic Image Dataset}
 \label{expt:synth}

\begin{figure*}[t]
\begin{minipage}[b]{0.18\linewidth}
\centerline{\epsfig{figure=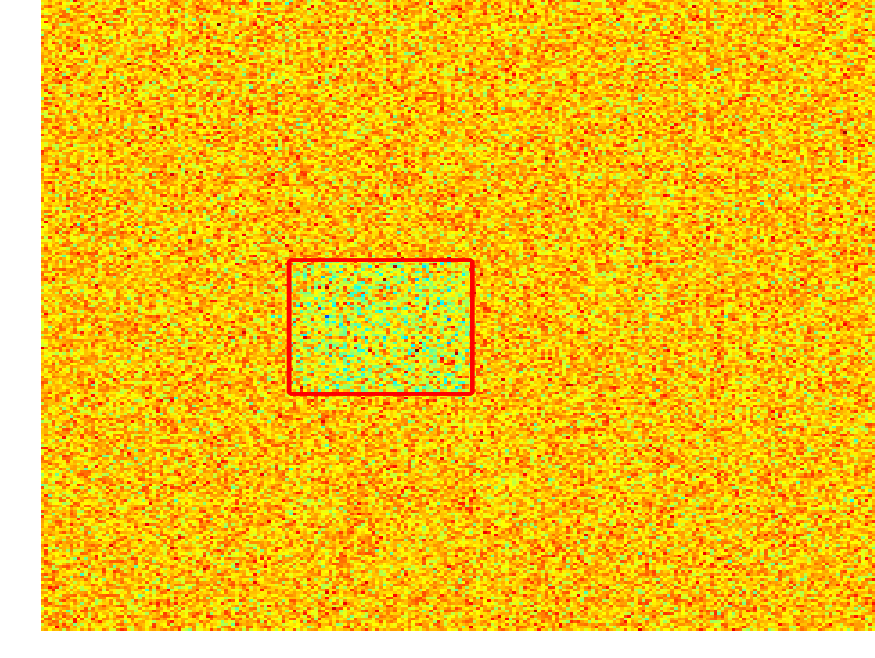,height=2.7cm,width=2.7cm}}  
\centerline{(a)}\medskip   \vspace{-0.25cm}
\end{minipage}
\hfill
\begin{minipage}[b]{0.18\linewidth}
\centerline{\epsfig{figure=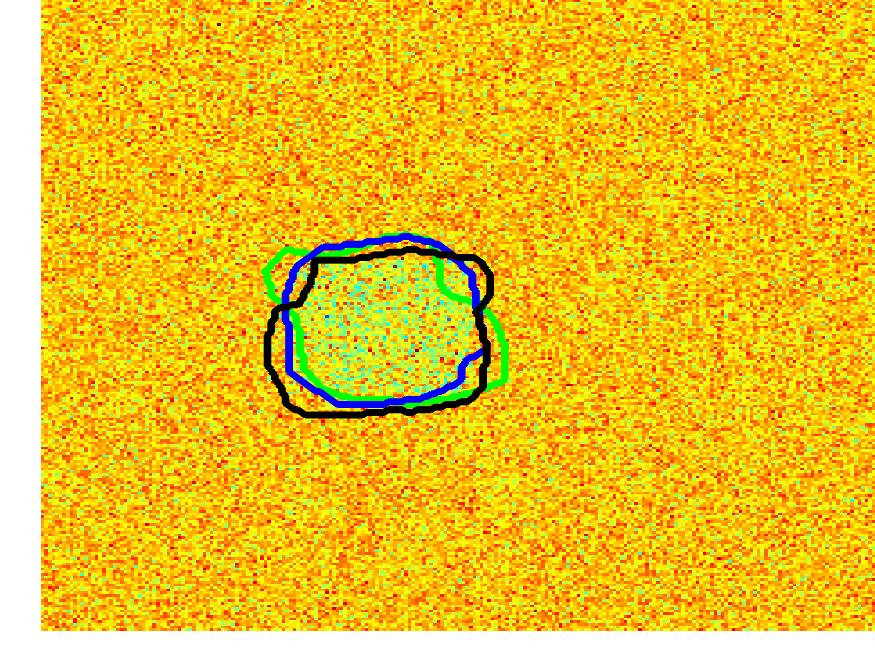,height=2.7cm,width=2.7cm}}  
\centerline{(b)}\medskip   \vspace{-0.25cm}
\end{minipage}
\hfill
\begin{minipage}[b]{0.18\linewidth}
\centerline{\epsfig{figure=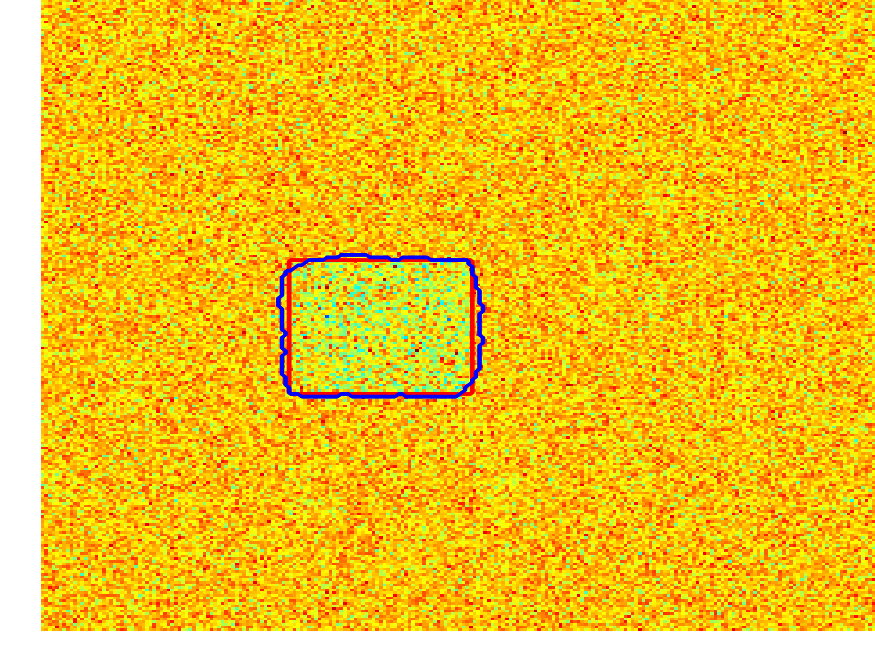,height=2.7cm,width=2.7cm}}  
\centerline{(c)}\medskip   \vspace{-0.25cm}
\end{minipage}
\hfill
\begin{minipage}[b]{0.18\linewidth}
\centerline{\epsfig{figure=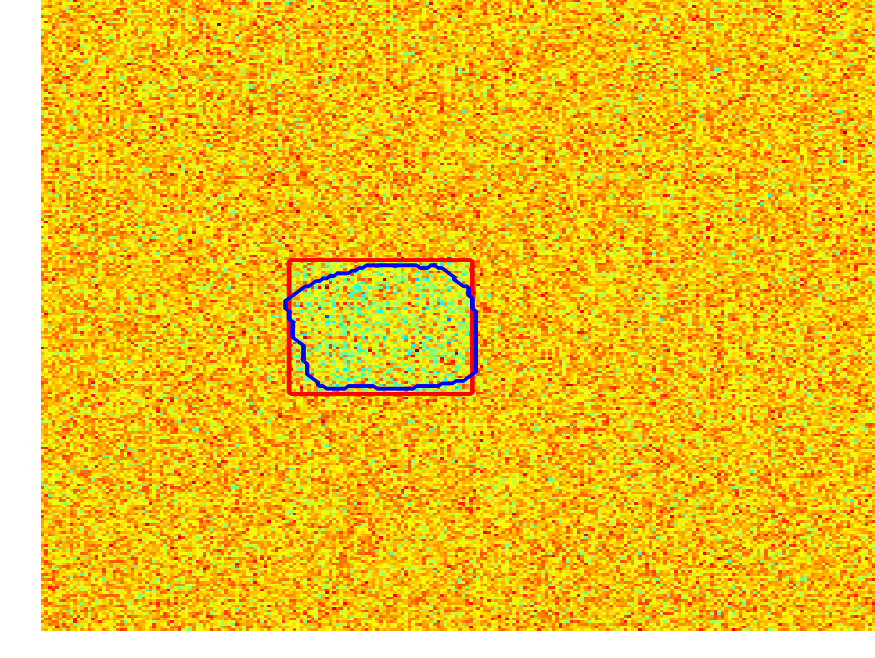,height=2.7cm,width=2.7cm}}  
\centerline{(d)}\medskip   \vspace{-0.25cm}
\end{minipage}
\hfill
\begin{minipage}[b]{0.18\linewidth}
\centerline{\epsfig{figure=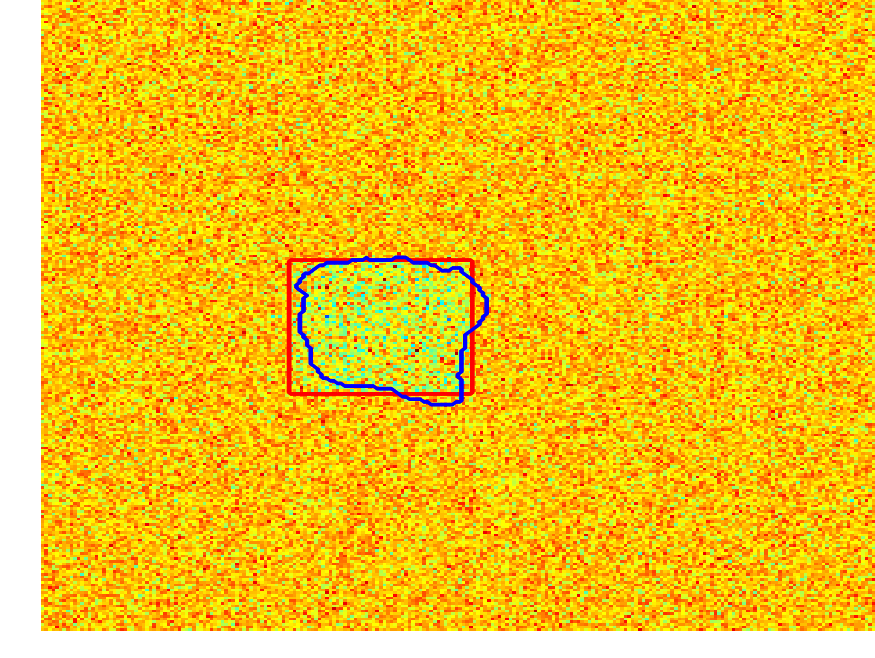,height=2.7cm,width=2.7cm}}  
\centerline{(e)}\medskip   \vspace{-0.25cm}
\end{minipage}
\hfill
\begin{minipage}[b]{0.18\linewidth}
\centerline{\epsfig{figure=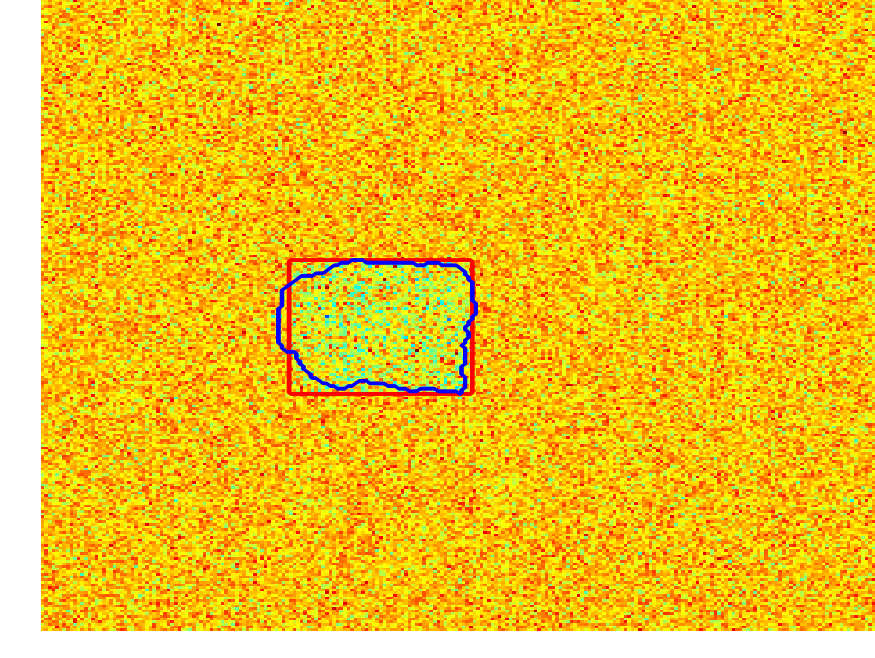,height=2.7cm,width=2.7cm}}  
\centerline{(f)}\medskip   \vspace{-0.25cm}
\end{minipage}
\hfill
\begin{minipage}[b]{0.18\linewidth}
\centerline{\epsfig{figure=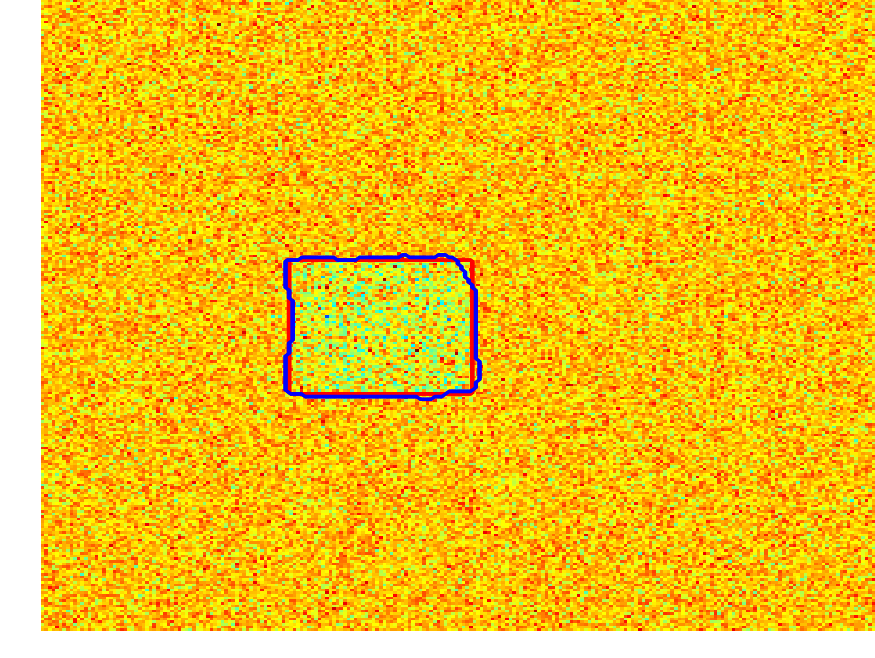,height=2.7cm,width=2.7cm}}  
\centerline{(g)}\medskip   \vspace{-0.25cm}
\end{minipage}
\hfill
\begin{minipage}[b]{0.18\linewidth}
\centerline{\epsfig{figure=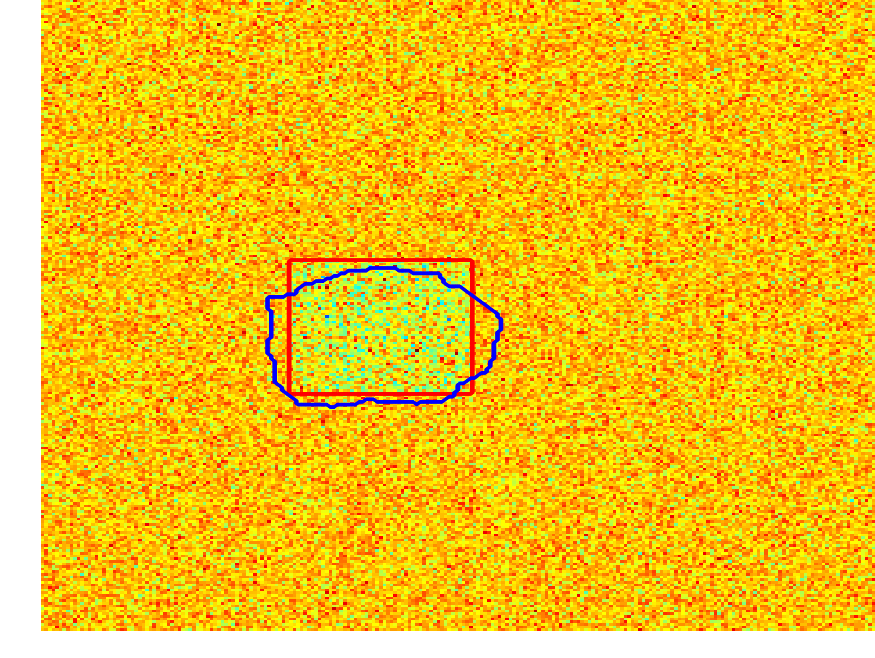,height=2.7cm,width=2.7cm}}  
\centerline{(h)}\medskip   \vspace{-0.25cm}
\end{minipage}
\hfill
\begin{minipage}[b]{0.18\linewidth}
\centerline{\epsfig{figure=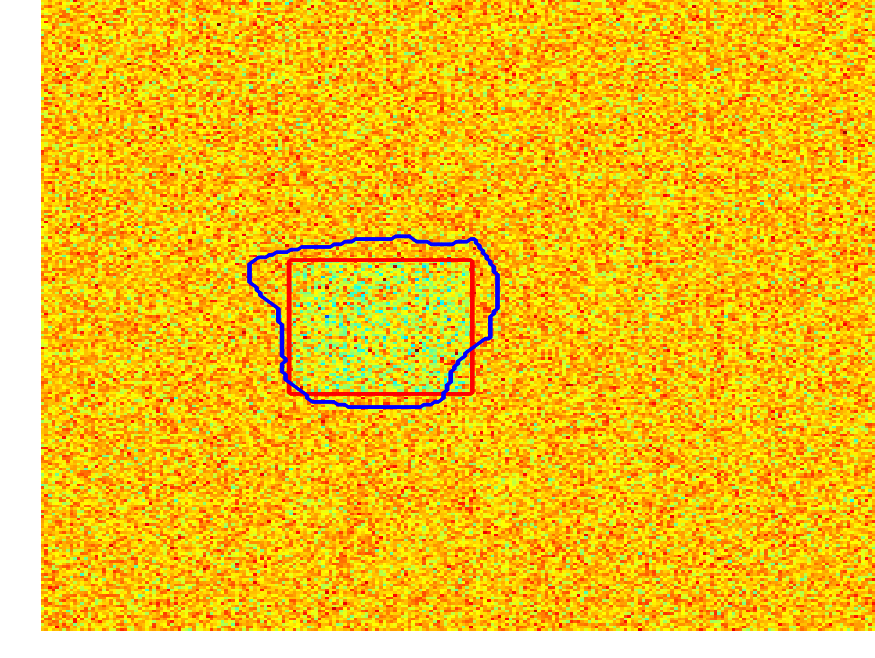,height=2.7cm,width=2.7cm}}  
\centerline{(i)}\medskip   \vspace{-0.25cm}
\end{minipage}
\hfill
\caption{(a) synthetic image with ground truth segmentation in red; (b) synthetic image with simulated expert annotations; final segmentation obtained by (c) $GC_{ME}$ (DM$=0.94$); (d) Majority voting (DM$=0.86$); (e) COLLATE (DM$=0.89$); (f) LMStaple (DM$=0.91$); (g) $GC_{ME-All}$ (DM$=0.96$); (h) $GC_{ME-wSSL}$ (DM$=0.83$); (i) $GC_{ME-wSC}$ (DM$=0.81$).}
\label{fig:Illus2}
\end{figure*}

  To illustrate the relevance of the SC score,
  we report segmentation results on synthetic images as they provide a certain degree of control on 
  image characteristics. 
  Figure~\ref{fig:Illus2} (a) shows an example synthetic image where the `diseased'
  region is within the red square. Pixel intensities are
  normalized to $[0,1]$. Intensities within the square
  have a normal distribution with $\mu\in[0.6,0.8]$
  and different $\sigma$. Background pixels have a lower 
  intensity distribution ($\mu\in[0.1,0.3]$ and different $\sigma$).
  $120$ such images with different shapes for the diseased region 
  (e.g., squares, circles, rectangles, polygons, of different dimensions)
  are created with known ground truths of the desired segmentation. 
   %
   %
   $15$ adjacent boundary points
  are chosen and randomly displaced between $\pm10-20$ pixels.
  This random displacement is repeated for $2-3$ more point sets depending on the size of the image.
  These multiple displacements of boundary points is the simulated annotation of one annotator.
  Two other sets of annotations are generated to create simulated annotations for $3$
  `experts'. 
  The annotations of different experts are shown as colored contours in Fig.~\ref{fig:Illus2} (b).

 To test our SSL based prediction strategy, we intentionally removed $1$ expert's annotations
  for each image/volume slice.
  The experts whose annotation was removed is chosen at random. 
  We refer to our method as $GC_{ME}$ (Graph Cut with Multiple Experts) and 
  compare its performance with  
 the final segmentations obtained using COLLATE \cite{COLLATE}, Majority Voting (MV) 
 \cite{LMS7}, and Local MAP-STAPLE (LMStaple) \cite{LocalMAPSTAPLE}.
  We also show results for $GC_{ME-All}$ in which none of the expert annotations
  were removed while predicting the final segmentation.
  Note that except for $GC_{ME-All}$, all other methods don't have access  
  to all annotations.

  Additionally, we show results for $GC_{ME-wSSL}$, i.e., $GC_{ME}$
  without SSL for predicting missing labels. In this case the penalty costs are 
  determined from $SC_i$'s of available annotations. Missing annotations of experts
  is not predicted and hence not used for determining the consensus segmentation. 
  Consensus segmentation results are also shown for $GC_{ME-wSC}$, i.e., $GC_{ME}$ without our SC score. 
  The penalty cost is the $\chi{2}$ distance between the reference
  distribution in the ground truth annotation of Fig.~\ref{fig:Illus2} (a),
  and the distribution from the `expert's' annotation. 
  Note that this condition can be tested only for synthetic images
  where we know the pixels' true labels.  
  For COLLATE we utilized the implementations available from the MASI fusion
package \cite{MASI}. Local MAP STAPLE 
implementation is available from the Computational Radiology
Laboratory website \cite{CRL}. For both methods we closely followed the 
parameter settings recommended by the authors.

  Table~\ref{tab:SynRes1} summarizes the performance of different methods.
  $GC_{ME-All}$ gives the highest DM and lowest HD values, followed by 
  $GC_{ME}$, \cite{LocalMAPSTAPLE}, \cite{COLLATE}, \cite{LMS7}, 
  $GC_{ME-wSSL}$ and $GC_{ME-wSC}$.
  Our proposed self consistency score accurately  
  quantifies the consistency level of each expert as is evident from the 
  significant difference in performance of $GC_{ME}$ and $GC_{ME-wSC}$ ($p<0.001$).
%
%
  Figures~\ref{fig:Illus2} (c)-(i) show the final segmentations obtained using the
  different methods. 

  %

  \begin{table*}[t]
	\centering
	\caption{Quantitative measures for segmentation accuracy on synthetic images. DM- Dice Metric in $\%$; HD is Hausdorff distance mm and $p$ is the result of Student $t-$tests with respect to $GC_{ME}$.}
\label{tab:SynRes1}
		\begin{tabular}{|l|l|l|l|l|l|l|l|}
		\hline
      {} & {$GC$}  & {$GC_{ME}$} & {LmStaple} & {Collate} & {MV} & {$GC$} & {$GC$} \\
      {} & {$_{ME-All}$}  & {} & { \cite{LocalMAPSTAPLE}} & {\cite{COLLATE}} & {\cite{LMS7}} & {$_{ME-wSSL}$} & {$_{ME-wSC}$} \\ \hline
      {DM} & {92.3} & {91.2} & {88.8} & {87.1} & {85.3} & {84.0} & {83.7} \\   \hline
      {HD} & {6.1} &  {7.4} & {9.0}  & {10.1} & {11.9} & {13.5} & {13.9} \\   \hline
      {$p$} & {0.032} & {-} & {$<0.01$} & {$<0.01$} & {$<0.01$} & {$<0.01$} & {$<0.001$} \\ \hline 
			%
	\end{tabular}
\end{table*}

%% file: MultipleExperts_Expts_Retina.tex
\subsection{Real Patient Retina Dataset}

  Quantitative evaluation is based on F-score and absolute pointwise localization error $B$ in pixels (measured in the radial direction). Additionally we report the overlap measure $S=Area(M\cap A)/Area(M\cup A)$. $M$ is the manual segmentation while $A$ is the algorithm segmentation.
 Comparative results are shown for $GC_{ME}$, $GC_{ME-All}$, $GC_{ME-wSSL}$, COLLATE, MV and LMStaple.

\begin{figure}[t]
\begin{minipage}[b]{0.14\linewidth}
\centerline{\epsfig{figure=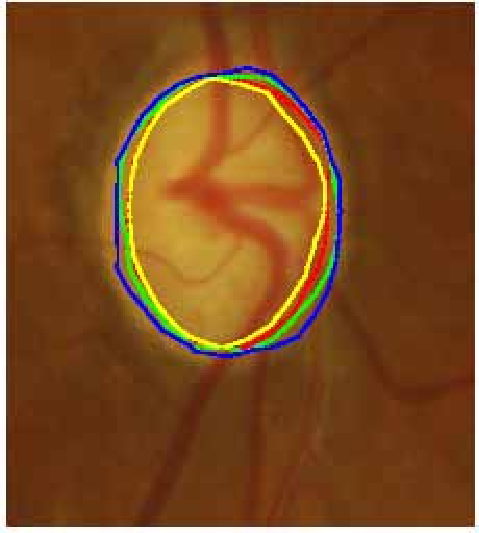,height=2.7cm,width=2.5cm}}  
\centerline{(a)}\medskip   \vspace{-0.25cm}
\end{minipage}
\hfill
\begin{minipage}[b]{0.142\linewidth}
\centerline{\epsfig{figure=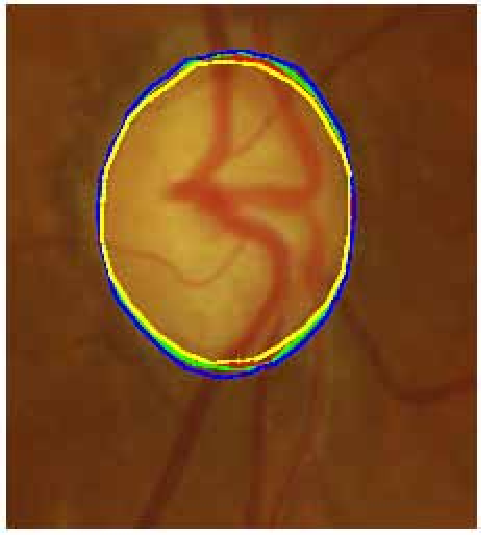,height=2.7cm,width=2.5cm}}  
\centerline{(b)}\medskip   \vspace{-0.25cm}
\end{minipage}
\hfill
\begin{minipage}[b]{0.142\linewidth}
\centerline{\epsfig{figure=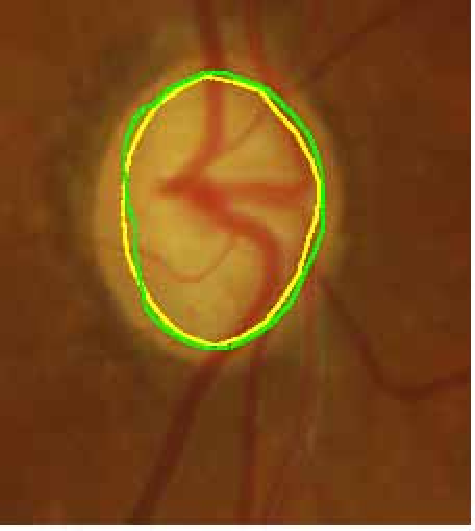,height=2.7cm,width=2.5cm}}  
\centerline{(c)}\medskip   \vspace{-0.25cm}
\end{minipage}
\hfill
%
\begin{minipage}[b]{0.142\linewidth}
\centerline{\epsfig{figure=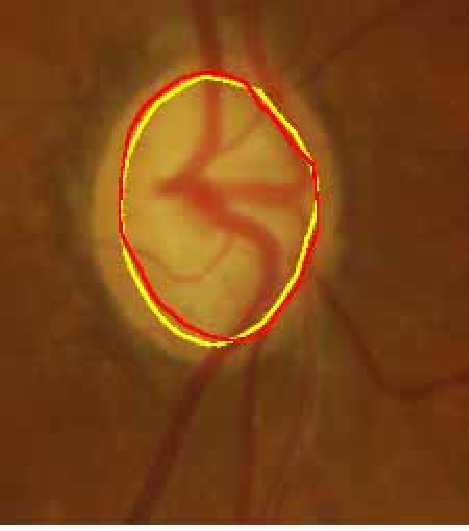,height=2.7cm,width=2.5cm}}  
\centerline{(d)}\medskip   \vspace{-0.25cm}
\end{minipage}
\hfill
\begin{minipage}[b]{0.142\linewidth}
\centerline{\epsfig{figure=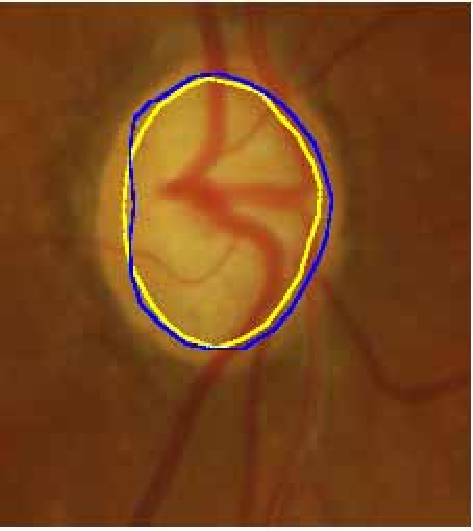,height=2.7cm,width=2.5cm}}  
\centerline{(e)}\medskip   \vspace{-0.25cm}
\end{minipage}
\hfill
\vspace{-0.4cm}
\caption{ Example annotations of (a) optic disc and (b) optic cup. The ground truth consensus segmentation is shown in yellow while the different expert annotations are shown in red, green and blue. Consensus segmentations for optic cup obtained using (c) $GC_{ME}$; (d) \cite{LocalMAPSTAPLE}; and (e) \cite{COLLATE}.}
\label{fig:GroundTruths}
\end{figure}

Table~\ref{tab:SegAcc} summarizes the segmentation performance of different methods. 
 %
 %
    %
 Figure~\ref{fig:GroundTruths} (b),(c) shows the individual expert annotations and the consensus ground truth annotation while Figs~\ref{fig:GroundTruths} (d)-(f) show the predicted ground truth for $3$
 fusion strategies. As is evident from the images $GC_{ME}$ shows the best agreement with the ground truth segmentations. 

  These results confirm our earlier observations from synthetic and CD patient datasets about: 1) the superior performance of $GC_{ME}$; 2) effectiveness of SSL in predicting missing annotation information;
  3) inferior performance of LMStaple due to predicting sensitivity and specificity 
   parameters from annotations without considering their overall consistency, and using EM; and 
   4) contribution of our SC score and graph cuts in obtaining better consensus annotations.

  \begin{table}[t]
\centering
\caption{Segmentation accuracy of retinal fundus images in terms of $F$ score, overlap and boundary distance for different methods. $B$ is in pixels; $Time$- fusion time in minutes;$F$-F score; $S$-overlap measure; $B$-boundary error.
}
\begin{tabular}{|l|l|l|l|l|l|l|}
\hline
 {} & {$GC$} & {$GC$}  & {COLLATE} & {LMStaple} & {$GC$} & {Majority }  \\ 
  {}  & {$_{ME}$} & {$_{ME-All}$}  & {\cite{COLLATE}} & { \cite{LocalMAPSTAPLE}} & {$_{ME-wSSL}$} & {Voting}  \\ \hline
 {F} &  {95.4} & {97.2} & {90.2} & {89.0} & {92.1}  & {86.4}   \\  \hline 
 {S} & {89.2} & {91.2}  & {84.8} & {83.2} & {85.9} & {80.8} \\  \hline  
 {B} & {9.9} & {8.2} & {13.2} & {10.9}  & {10.3} & {18.1}    \\ \hline
{Time}  & {7} & {7} & {6} & {9} & {7} & {3}  \\ \hline 
\end{tabular}
\label{tab:SegAcc}
\end{table}

   \begin{figure}[t]
\begin{minipage}[b]{0.152\linewidth}
\centerline{\epsfig{figure=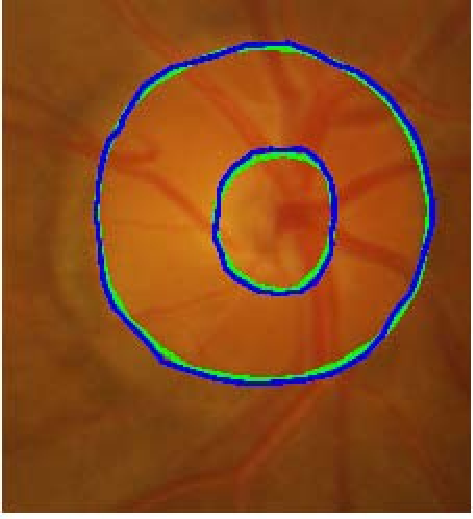,height=2.6cm,width=2.6cm}}  
\centerline{(a)}\medskip   \vspace{-0.25cm}
\end{minipage}
\hfill
\begin{minipage}[b]{0.152\linewidth}
\centerline{\epsfig{figure=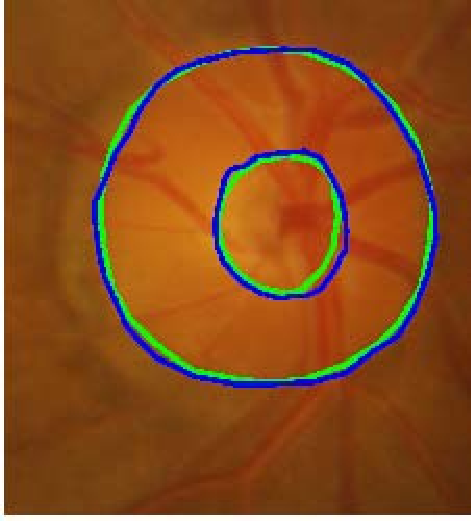,height=2.6cm,width=2.6cm}}  
\centerline{(b)}\medskip   \vspace{-0.25cm}
\end{minipage}
\hfill
\begin{minipage}[b]{0.152\linewidth}
\centerline{\epsfig{figure=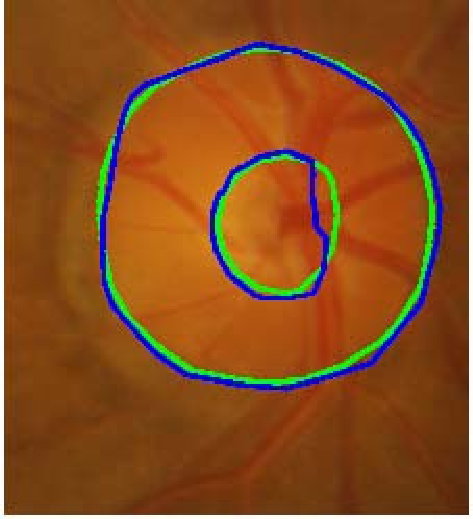,height=2.6cm,width=2.6cm}}  
\centerline{(c)}\medskip   \vspace{-0.25cm}
\end{minipage}
\hfill
\begin{minipage}[b]{0.15\linewidth}
\centerline{\epsfig{figure=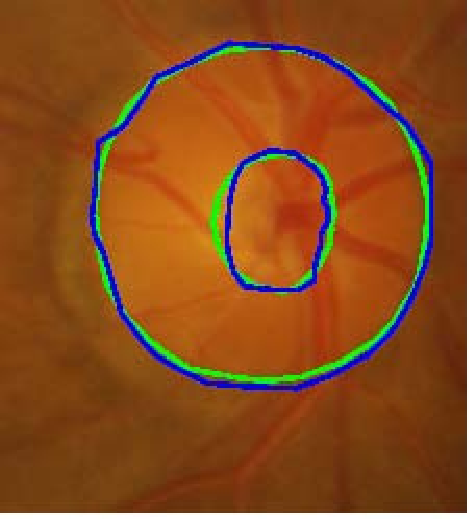,height=2.6cm,width=2.6cm}}  
\centerline{(d)}\medskip   \vspace{-0.25cm}
\end{minipage}
\hfill
\begin{minipage}[b]{0.152\linewidth}
\centerline{\epsfig{figure=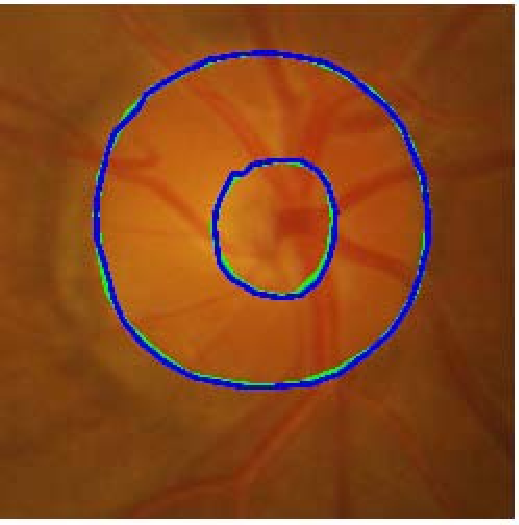,height=2.6cm,width=2.6cm}}  
\centerline{(e)}\medskip   \vspace{-0.25cm}
\end{minipage}
\hfill
\vspace{-0.4cm}
\caption{Segmentation results for different methods: (a) our proposed $GC_{ME}$ method (b) \cite{LocalMAPSTAPLE}; (c) \cite{COLLATE}; (d) Majority Voting; and (e) $GC_{ME-All}$. Green contour is manual segmentation and blue contours are algorithm segmentations from different fusion methods. }
\label{fig:SegRes}
\end{figure}

%% file: MultipleExperts_Concl.tex
\section{Discussion And Conclusion}
\label{sec:disc}

We have proposed a novel strategy for combining multiple
annotations and applied it for segmenting Crohns disease tissues from abdominal MRI, and the optic cup and disc from retinal fundus images.
 Qualitative evaluation is performed using a machine learning approach for segmentation.
 Highest segmentation accuracy is observed for the annotations obtained
 by our fusion strategy, which is indicative of better quality annotations.
  The comparative results of our method and other fusion strategies highlight the following
  major points.
  \begin{enumerate}
  \item With least variance of DM values $GC_{ME}$ is the most consistent fusion method, and with highest DM values is also the  most accurate.

    \item  SSL effectively predicts missing annotation information since $GC_{ME}$
  has very close performance to $GC_{ME-All}$, 
  and is significantly better than $GC_{ME-wSSL}$.
  Local MAP STAPLE infers missing annotations
  by minimising the log-likelhood of the overall cost function. Employing EM contributes to its  erroneous results.
   SSL's advantage is the predicted annotations are consistent with 
   previously annotated samples by considering both global information and local feature consistencies. 
  
  \item Our proposed self consistency score accurately  
  quantifies the consistency level of each expert as is evident from the 
  performance of $GC_{ME}$ and $GC_{ME-wSC}$ ($p<0.001$) for synthetic images.
  SC analyses feature distributions of neighboring pixels
  that share the same labels and gives higher values for  consistent annotations which have 
  similar feature distributions.
 
 \item Graph cut optimization produces a quick global optimum without
 the risk of getting trapped in local minimum which can be a serious limitation for 
 EM based methods. Use of GC and SSL together contribute to low computation time
 since there is no iterative approach involved.

  
  \end{enumerate}
  
  \dmr{
	Our proposed method for obtaining consensus annotations can be used in scenarios where there is the need to find a ground truth. In most medical image analysis applications it is good practice to have $2$ or more experts annotate the images. This also minimizes scope of biased or inaccurate annotations. In such cases our method can be used to generate the ground truth from the multiple expert annotations. However, in reality it can be difficult to obtain multiple expert annotations due to cost and resource issues. In such scenarios multiple segmentations can be generated from different automatic segmentation algorithms and the consensus ground truth segmentation can be generated using our method.
	}
  
 \dmr{
\textbf{Algorithm Limitations}:
	It is important that in order to generate a good ground truth we have multiple experts' annotations. As mentioned before that it is not easy for many experts to provide annotations. Although in principle we can use different segmentation algorithms to generate candidate segmentations and then calculate the ground truth, these algorithms may not always be accurate and the final result would be erroneous. Thus we see that our algorithm's performance is limited by the availability of qualified experts to provide accurate annotations.
	}
	
	\dmr{
  SSL for predicting missing annotations is 
 an important part of our fusion approach and erroneous prediction affects the final results.
 In SSL the unlabeled samples are assigned a class based on their presence in the feature space and 
 its subsequent split to maximize information gain. Erroneous labels of one or more annotations 
 affects the predicted label. However, our proposed method limits the damage due to 
 inaccurate label predictions with the help of the SC score which is based 
 on the image features of each annotation. Inaccurately labeled annotations are assigned low scores 
 since the image features for a label are not consistent throughout the annotation. Subsequently, 
 inaccurate or ambiguous annotations have a lower contribution to the final consensus segmentation.
 Although we cannot completely eliminate mistakes, use of SC allows us to minimize them by 
 assigning lower importance to erroneous annotations.
}